\documentclass[10pt,fleqn]{article}
\usepackage [top=25truemm,bottom=25truemm,left=25truemm,right=25truemm]{geometry}

\newcommand{\narrowbaselines}{\relax}
\usepackage{tex_elements/tysty}
\usepackage{epstopdf}
\usepackage{subcaption}
\newcommand{\refsubref}[2]{\ref{#1}(\subref{#1_#2})}

\def\CDR{\mathrm{CDR}}

\newcommand{\stp}[1]{{\bsf [STEP #1]}}
\newcounter{stepnum} 
\newenvironment{step}[1]{\quad\\ \noindent {\stp{\refstepcounter{stepnum}\thestepnum} {\bsf #1}}\\}{\quad\\}
\newcommand{\case}[1]{{\bsf [CASE #1]}}
\newcounter{casenum} 

\newcommand{\orcidicon}{\includegraphics[width=0.32cm]{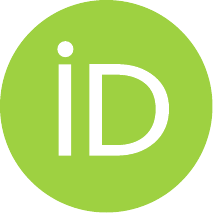}}
\foreach \x in {A,..., Z}{%
\expandafter\xdef\csname orcid\x\endcsname{\noexpand\href{https://orcid.org/\csname orcidauthor\x\endcsname}{\noexpand\orcidicon}}
}

\title{\textbf{Autocorrelation effects in a stochastic-process model\\
for solving two-armed bandit problems}}
\author{Tomoki Yamagami{}$^{\,1,\,*}$ \orcidA
\and Mikio Hasegawa{}$^{\,2}$ \orcidB
\and Takatomo Mihana{}$^{\,3}$ \orcidC
\and Ryoichi Horisaki{}$^{\,3}$ \orcidD
\and Atsushi Uchida{}$^{\,1,\,\dagger}$ \orcidE
}
\date{}

\begin{document}
\columnseprule=0.2mm
\maketitle
\vspace{-2.1\baselineskip}
\begin{center}
{\small
$^1$ Department of Information and Computer Sciences, Saitama University,\\
255 Shimo-Okubo, Sakura-ku, Saitama City, Saitama 338--8570, Japan.\\
$^2$ Department of Electrical Engineering, Tokyo University of Science,\\
6--3--1, Niijuku, Katsushika-ku, Tokyo 125--8585, Japan.\\
$^3$ Department of Information Physics and Computing, The University of Tokyo,\\
7--3--1 Hongo, Bunkyo-ku, Tokyo 113--8656, Japan.\\
$^*$\texttt{tyamagami@mail.saitama-u.ac.jp}\quad $^\dagger$\texttt{auchida@mail.saitama-u.ac.jp}
}\vspace{1\baselineskip}\\
\end{center}

\begin{abst}
Decision makers exploiting photonic chaotic dynamics obtained by semiconductor lasers provide an ultrafast approach to solving multi-armed bandit problems by using a temporal optical signal as the driving source for sequential decisions. 
In such systems, the sampling interval of the chaotic waveform shapes the temporal correlation of the resulting time series, and experiments have reported that decision accuracy depends strongly on this autocorrelation property.
However, it remains unclear whether the benefit of autocorrelation can be explained by a minimal mathematical model.
Here, we analyze a stochastic-process model for solving the two-armed bandit problem based on time series, where the threshold and a two-valued Markov signal evolve jointly.
Numerical results reveal an environment-dependent structure: negative (positive) autocorrelation is optimal in reward-rich (reward-poor) environments. 
These findings show that negative autocorrelation of the time series is advantageous 
when the sum of the winning probabilities is more than one, whereas positive autocorrelation is useful when the sum of the winning probabilities is less than one.
Moreover, the performance is independent of autocorrelation if the sum of the winning probabilities equals one, which is mathematically clarified.
This study paves the way for solving the two-armed bandit problems
for reinforcement learning applications in wireless communications and robotics.
\end{abst}


\setcounter{equation}{0}
\section{Introduction}\label{introduction}
The rapid progress of artificial intelligence and machine learning has stimulated research into optics- and photonics-based computing systems \cite{jahns2014optical, kitayama2019novel, shastri2021photonics}.
One promising direction is photonic decision making, in which physical properties of light are exploited to solve multi-armed bandit problems \cite{robbins1952some}.
The multi-armed bandit problem represents one of the simplest settings of reinforcement learning. Reinforcement learning is a framework for solving tasks through interaction between a decision maker (an agent) and the object of its action (an environment). After the agent takes an action, the environment probabilistically returns a reward as feedback, and the goal of reinforcement learning is to maximize the cumulative reward over a sequence of actions. In general, the agent does not know in advance which behavior is more effective for obtaining higher rewards; thus, the agent must explore different actions and learn how to optimize its behavior. Moreover, the best choice found by the agent may not always remain optimal because the scenario can change depending on external conditions and the agent's action history, which is modeled as states.

In contrast, the multi-armed bandit problem assumes a simple setting with respect to the state: the optimal choice does not change, although it is hidden from the agent. Specifically, the multi-armed bandit problem considers repeated selections among multiple slot machines (arms) in the environment. Each selection of an arm corresponds to an action in the reinforcement-learning framework, and the selected arm yields a reward with an arm-dependent probability, which we call the winning probability of the arm. The agent aims to maximize the cumulative reward through a series of arm selections.
A central challenge in the multi-armed bandit problem is the exploration--exploitation trade-off \cite{march1991exploration,daw2006cortical}. 
As described above, the agent needs to try each possible selection multiple times to learn which arm should be chosen to optimize the total reward. On the other hand, the number of selections is limited in general, meaning that excessive exploration can waste opportunities to choose the arm that is expected to provide the best outcome. Therefore, balancing exploration and exploitation is crucial in the multi-armed bandit problem. To achieve this balance, various algorithms have been developed, such as the $\varepsilon$-greedy \cite{sutton2018reinforcement}, softmax \cite{sutton2018reinforcement, daw2006cortical}, upper confidence bound \cite{auer2002finite}, and quantum walks~\cite{yamagami2023bandit}.
Given this broad applicability and the need for efficient decision making under uncertainty, multi-armed bandit problems have also attracted sustained attention as a practical framework.
Originally formulated in the context of sequential allocation in clinical trials \cite{thompson1933likelihood}, multi-armed bandit problems are now employed in applications such as online advertising \cite{nakamura2005improvements} and wireless communications \cite{lai2008medium}.

Photonic implementations based on semiconductor lasers, referred to as the laser-chaos-based decision maker, have attracted significant attention for solving the multi-armed bandit problem~\cite{naruse2017ultrafast}.
This system utilizes chaotic time series generated by a semiconductor laser with optical feedback \cite{ohtsubo2012semiconductor,soriano2013complex} to physically implement the decision-making process.
Chaotic laser signals provide quasi-irregular oscillations, and chaotic laser dynamics have been studied for secure key distribution \cite{colet1994digital,argyris2005chaos,scheuer2006giant,yoshimura2012secure}, random number generation \cite{uchida2008fast,argyris2010implementation,kanter2010optical}, and reservoir computing \cite{appeltant2011information,duport2012all,kanno2020adaptive}.
In the context of decision making, the chaotic signal is fed to a decision-making scheme established on the tug-of-war (ToW) principle \cite{kim2010tug,kim2015efficient}.
This algorithm, inspired by slime-mold behavior, dynamically adjusts a threshold to regulate arm-selection frequencies in two-armed bandit environments.
In this scheme, the signal amplitude is binarized by comparison with the threshold, corresponding to which arm is selected, and ultrafast (GHz-order) decision-update rates have been demonstrated \cite{naruse2017ultrafast}.
Building on this baseline, later studies have extended photonic bandit solvers toward larger-scale and more practical settings by harnessing additional physical degrees of freedom and control parameters \cite{naruse2018scalable,homma2019chip, peng2021photonic, iwami2022controlling, morijiri2023parallel, shen2023harnessing, cuevas2024solving}.
This ultrafast adaptive property is also a crucial resource for the research fields that demand rapid decision making such as wireless communication \cite{takeuchi2020dynamic}, real-time model selection \cite{kanno2020adaptive}, and Ising-machine optimization \cite{yasudo2025photonic}.

A key empirical finding is that the autocorrelation properties of the chaotic signal have a strong influence on decision-making accuracy: sampling the signal at delay times that induce negative autocorrelation reliably improves performance compared with time series lacking such autocorrelation \cite{naruse2017ultrafast}.
This improvement has also been confirmed using Fourier surrogate data that preserve autocorrelation while removing higher-order temporal dependencies \cite{okada2021analysis}.
Related motivations appear in other contexts as well, such as code-division multiple access (CDMA), where minimizing correlation is essential for improving performance \cite{kohda2003pursley, hasegawa2016improving}.

To advance theoretical understanding of autocorrelation-driven performance, a recent work has introduced a stochastic-process model that reproduces the time evolution of decision making based on comparing time series and threshold, as in the laser-chaos-based decision maker \cite{okada2022theory}.
In this model, the threshold is regarded as the position of a random walker whose transition probabilities depend on the sign of the time series; the signal values are governed by a two-valued Markov chain.
The prior study \cite{okada2022theory} reports that the model demonstrates how negative autocorrelation can enhance performance under specific environmental conditions.
However, the evidence is limited to only three cases regarding the pair of the winning probabilities of two arms, which is insufficient to conclude that negative autocorrelation universally improves decision-making performance.

In this study, we analyze a stochastic-process model to clarify the relationship between the autocorrelation of the driving signal and decision-making performance across a broad range of environments.
We first provide an overview of the decision-making scheme based on a comparison between the time series and the adjustable threshold, named the \textit{time-series-based decision making} in this paper.
Note that the laser-chaos-based decision maker adopts the time-series-based decision making, in which the time series is obtained by the chaotic laser dynamics.
We present the stochastic-process model describing the evolution of the time series and threshold dynamics in the time-series-based decision making, where the signal is defined by a two-valued Markov process.
We present numerical results for the model
and offer discussions of our findings and future prospects.
\setcounter{equation}{0}
\section{Modeling}\label{modeling}

In this section, we first explain the time-series-based decision making, which is the decision-making scheme conducted by comparing a time series and a threshold for the two-armed bandit problem between arms $A$ and $B$.
Then, we focus on the case that the time series is defined by the two-valued Markov chain and mention the stochastic-process model to elaborate the time variation of the signal and threshold.

We denote the winning probabilities of arms $A$ and $B$ by $p_A$ and $p_B$, respectively.
Throughout this study, we assume $p_A > p_B$, meaning that arm $A$ is the optimal arm.

\subsection{Time-series-based decision making}\label{modeling:subsec:signal}

\begin{figure}[ht]
    \centering
	\begin{minipage}[t]{0.48\textwidth}
		\includegraphics[width=\textwidth]{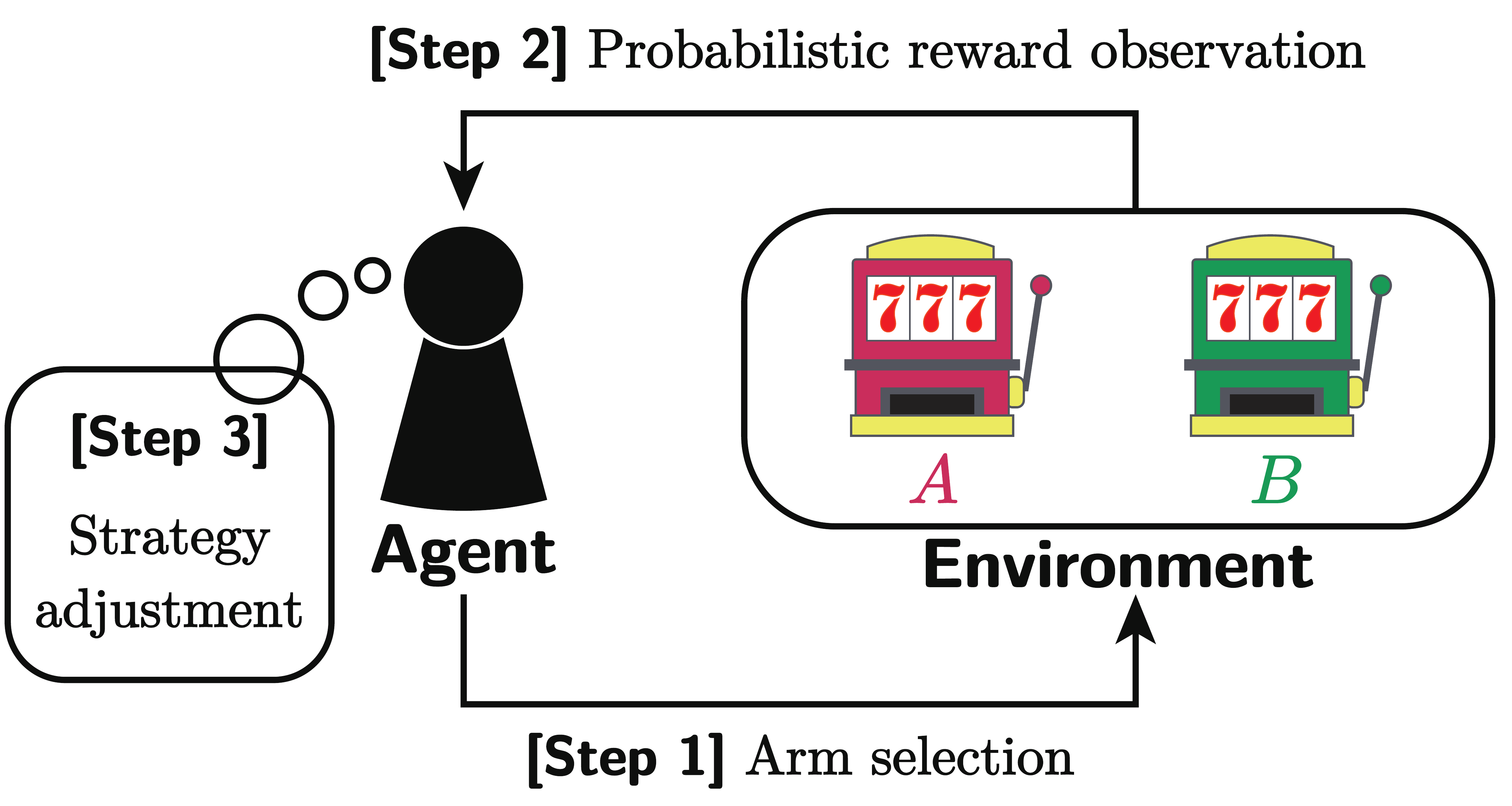}
		\subcaption{}\label{modeling:fig:mabtow_mab}
	\end{minipage}\hfill
	\begin{minipage}[t]{0.48\textwidth}
		\includegraphics[width=\textwidth]{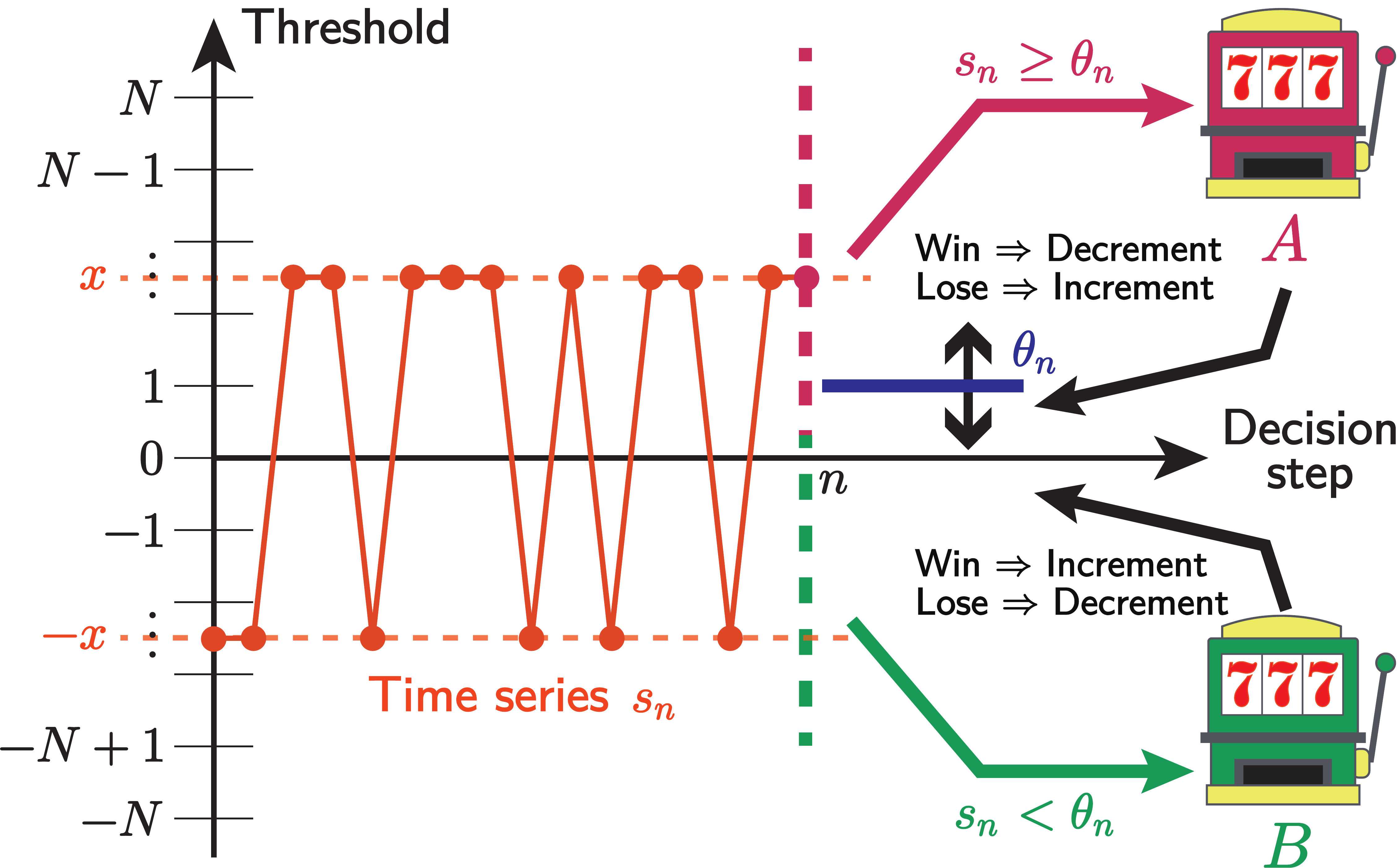}
		\subcaption{}\label{modeling:fig:mabtow_schematic}
	\end{minipage}
    \caption{Schematic of the two-armed bandit problem driven by a time series.
    \textbf{(a)} 
    Three steps in a single decision-making cycle.
    The agent selects an arm and probabilistically observes a reward from the selected arm.
    Based on the outcome, the agent adjusts its strategy (i.e., the criterion for arm selection), which in turn determines the next selection.
    Multi-armed bandit problems describe optimization through repetition of these three steps.
    \textbf{(b)} 
    Schematic of time-series-based decision making by comparison between the time series and the threshold.
    At the $n$-th time step, the decision is made by comparing the given signal value $s_n$ with the adjustable threshold $\theta_n$.
    If $s_n$ is greater than $\theta_n$, the agent selects arm $A$; otherwise, it selects arm $B$.
    When the selected arm wins (i.e., generates a reward), the threshold is updated so that the same selection becomes more likely;
    otherwise, it is updated so that the other selection becomes more likely.
    In the laser-chaos-based decision maker, the signal $s_n$ is obtained by sampling chaotic feedback, where $n$ also denotes the sampling index of the chaotic time series~\cite{naruse2017ultrafast}.
    In this paper, we assume that the time series $s_n$ is generated by the two-valued Markov chain for simplicity.
    }\label{modeling:fig:mabtow}
\end{figure}

The time-series-based decision making operates according to the following three-step procedure:  
{\bsf [STEP 1]} Arm selection,  
{\bsf [STEP 2]} Probabilistic reward observation, and  
{\bsf [STEP 3]} Strategy adjustment,
consistent with the standard formulation of the multi-armed bandit problem shown in Fig.~\refsubref{modeling:fig:mabtow}{mab}.  
Iterating these steps describes the interaction between the agent and the environment, corresponding to a reinforcement learning process.  
The decision maker uses a given time series $s_n$ and an adjustable threshold value $\theta_n$ to make the $n$-th decision.

\begin{step}{Arm selection}
The agent selects arm $A$ or $B$ based on the comparison between the instantaneous signal value $s_n$ and the threshold $\theta_n$.  
If $s_n \ge \theta_n$, the agent selects arm $A$; otherwise, arm $B$ is selected, as shown in Fig.~\refsubref{modeling:fig:mabtow}{schematic}.
\end{step}
\begin{step}{Probabilistic reward observation}
The selected arm $a_n \in \{A,\,B\}$ yields a reward $r_n$ drawn from a Bernoulli distribution $\operatorname{Ber}(p_{a_n})$:
\begin{equation}
	r_n =
	\begin{cases}
		1 & \text{with probability } p_{a_n},\\
		0 & \text{with probability } 1-p_{a_n}.
	\end{cases}
\end{equation}
Thus, the agent observes a binary reward governed by the winning probability of the selected arm.
\end{step}
\begin{step}{Strategy adjustment}
Based on the outcome obtained in {\bsf [STEP 2]}, the agent updates the threshold value $\theta_{n+1}$ for the next decision.  
The threshold is modified by unit increments or decrements; however, it is constrained within the fixed bounds $\pm N$ for a given natural number $N$.  
If the update attempts to push the threshold beyond these bounds, it remains clamped at the boundary.

The update rule is given as follows. Under selection of arm $A$,
\begin{equation}\label{modeling:eq:updateA}
	\theta_{n+1} =
	\begin{cases}
		\max\{\theta_n - 1,\,-N\} & \text{if arm $A$ wins},\\
		\min\{\theta_n + 1,\, N\} & \text{if arm $A$ loses},
	\end{cases}
\end{equation}
and under selection of arm $B$,
\begin{equation}\label{modeling:eq:updateB}
	\theta_{n+1} =
	\begin{cases}
		\min\{\theta_n + 1,\,N\} & \text{if arm $B$ wins},\\
		\max\{\theta_n - 1,\,-N\} & \text{if arm $B$ loses}.
	\end{cases}
\end{equation}

This rule is designed to increase the likelihood of making better decisions in subsequent rounds.  
When arm $A$ wins, the agent is encouraged to choose arm~$A$ again; decreasing the threshold makes this more likely because $s_n$ needs to only exceed a smaller value.  
Conversely, if arm $A$ loses, increasing the threshold encourages the agent to switch to arm~$B$.  
The rule for arm~$B$ follows the same rationale.
\end{step}

The present threshold dynamics can be regarded as a simplified case of the laser-chaos-based decision maker \cite{naruse2017ultrafast} with the memory parameter $\alpha = 1$ (see Eqs.~\eqref{methods:eq:TAA} and \eqref{methods:eq:TAB} in the \textit{Methods} section).
Details on the correspondence between the time-series-based decision making and the laser-chaos-based decision making and its parameters are provided in the \textit{Methods} section.

\subsection{Stochastic process model}\label{modeling:subsec:stochastic}
We introduce a stochastic process model to formalize the time-series-based decision-making process described above.  
The model is based on the previous work~\cite{okada2022theory}, with modifications to notation and a generalization of the parameter $x$ defining the signal series $s_n$.

We assume that the time series $(s_n)_{n\in\mathbb{N}}$, indexed by the set of positive integers $\mathbb{N}$, takes one of two values, $x$ or $-x$ with $x \in (0,\,N)\setminus\mathbb{N}$,
where integer values of $s_n$ are excluded for simplicity.
This generalizes the previous work~\cite{okada2022theory}, in which $x$ was restricted to $(N-1,\,N)$.  

The series $(s_n)$ is modeled as a two-valued Markov chain, with switching probability $\gamma$:
\begin{equation}\label{stochastic:eq:gamma}
 \mathbb{P}(s_{n+1}=\mp x \mid s_n=\pm x) = \gamma, \qquad
 \mathbb{P}(s_{n+1}=\pm x \mid s_n=\pm x) = 1-\gamma,
\end{equation}
where $\mathbb{P}(E|F)$ denotes the conditional probability with which an event $E$ occurs under the condition that another event $F$ is true.
Let $\lambda$ be the autocorrelation coefficient of $(s_n)$, and the following holds:
\begin{equation}\label{modeling:eq:lambda}
 \lambda = \fraction{\mathbb{E}[s_n s_{n+1}]}{\mathbb{E}[s_n^2]} = 1 - 2\gamma, \quad \text{hence}\quad \gamma = \frac{1-\lambda}{2},
\end{equation}
where $\mathbb{E}[U]$ denotes the expectation of the random variable $U$.
It is worth noting that negative autocorrelation ($\lambda < 0$) corresponds to $\gamma > 1/2$, indicating that $s_n$ is more likely to switch between $x$ and $-x$.

It should be noted that the term ``two-valued'' applies only to the signal process $(s_n)$, not to the whole state of the model.
Here, $x$ determines the two signal levels $\pm x$, whereas $N$ determines the bounded integer range of the threshold $\theta_n$ introduced in the previous section.
For instance, the setting $N=4$ and $x=3.5$ corresponds to $s_n\in\{-3.5,\,3.5\}$ and $\theta_n\in\{-4,-3,\ldots,3,4\}$.

\begin{figure}[ht]
    \centering
	\begin{minipage}[t]{0.48\textwidth}
		\includegraphics[width=\textwidth]{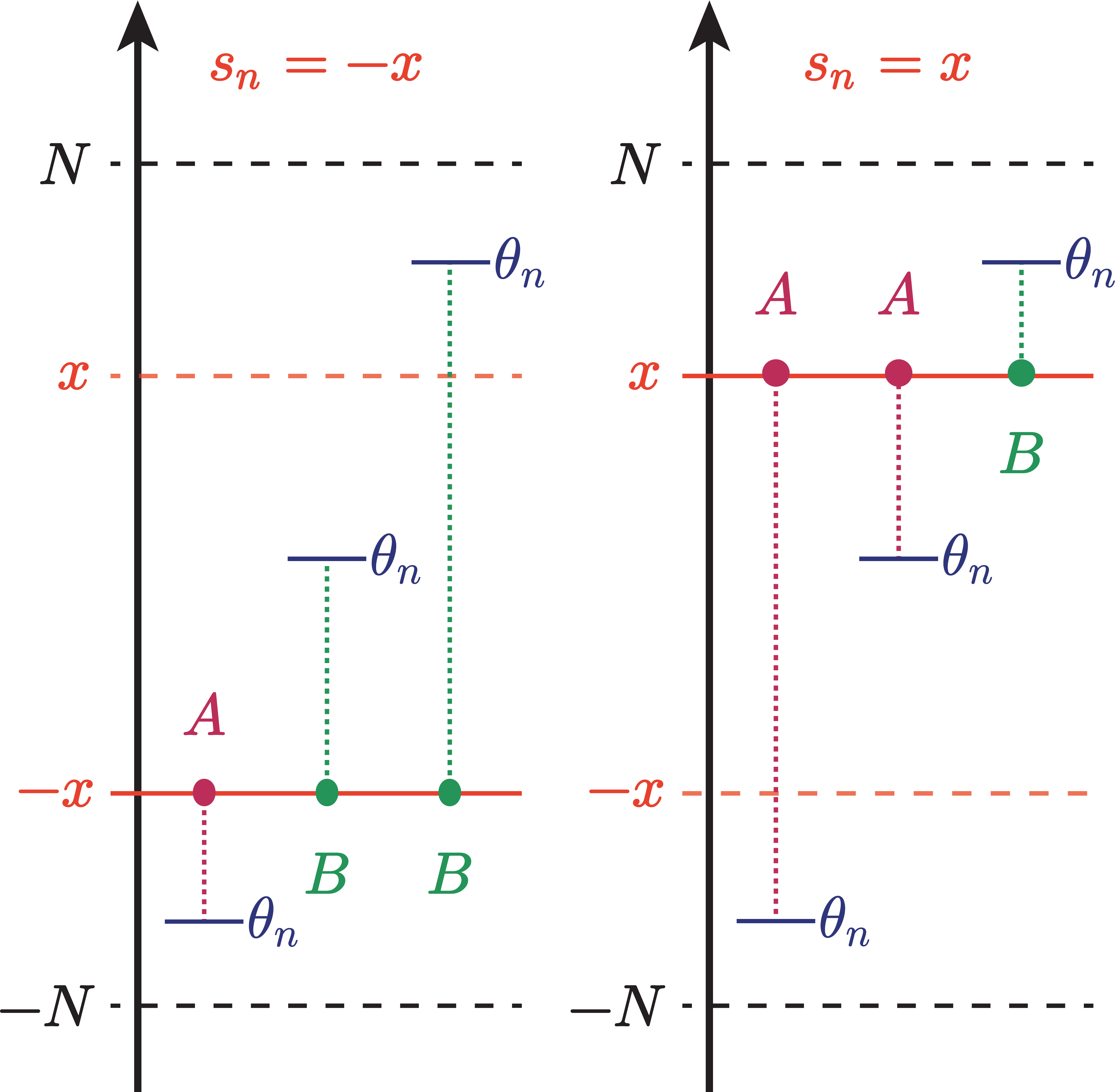}
		\subcaption{}\label{modeling:fig:towsn_negaposi}
	\end{minipage}\hfill
	\begin{minipage}[t]{0.48\textwidth}
		\includegraphics[width=\textwidth]{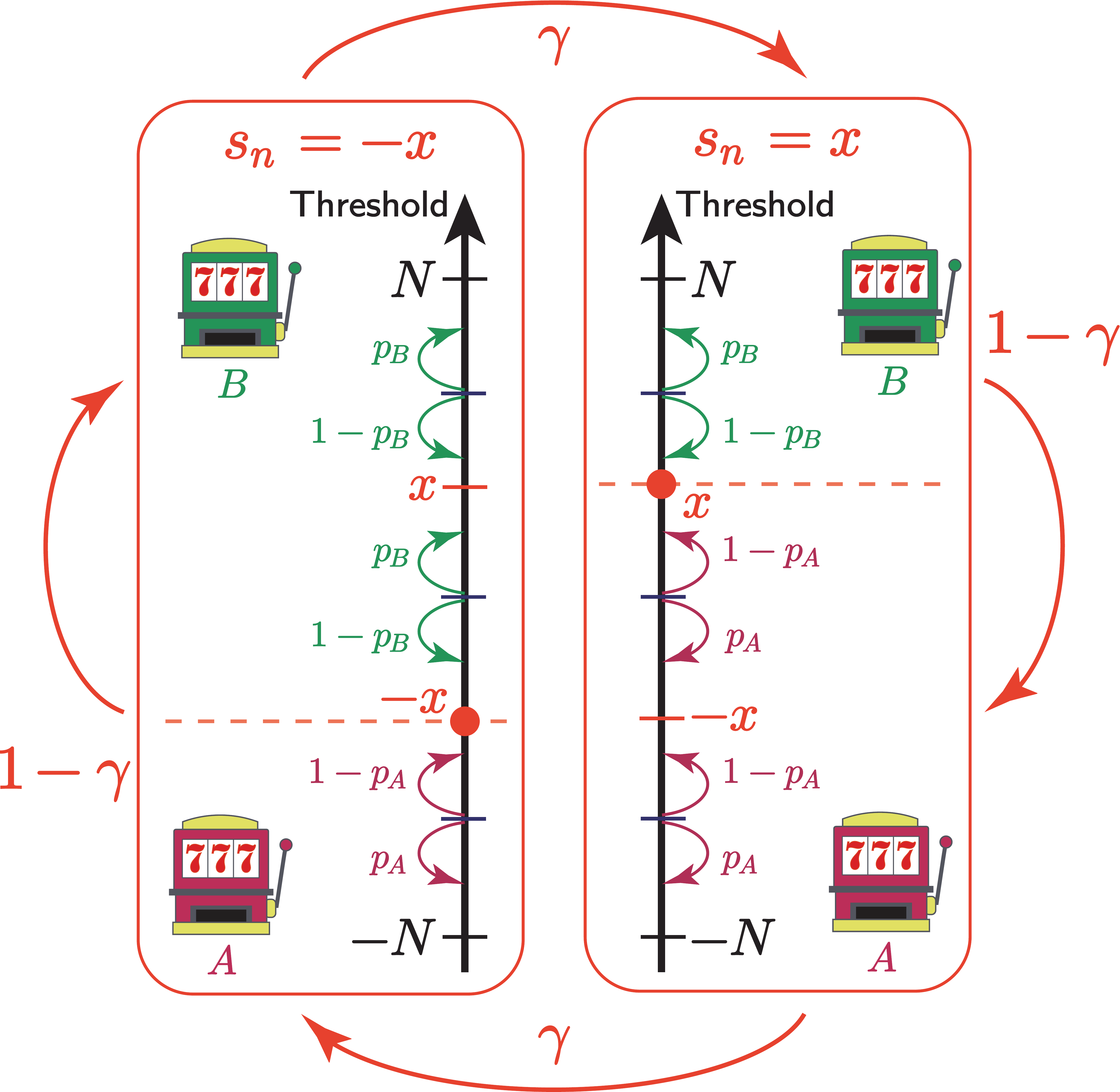}
		\subcaption{}\label{modeling:fig:towsn_rw}
	\end{minipage}
    \caption{Schematic on the two-armed bandit problem driven by the two-valued Markov chain as the time series $s_n$.
    \textbf{(a)} 
    The relation between the threshold $\theta_n$ and which arm $A$ or $B$ is selected for the time series $s_n = -x$ (left panel) and $s_n = x$ (right panel). 
    For $s_n = -x$, arm $A$ is selected under the condition $\theta_n < -x$; otherwise, arm $B$ is selected.
    For $s_n = x$, arm $B$ is selected under the condition $\theta_n > x$; otherwise, arm $A$ is selected.
    \textbf{(b)} 
    State-transition (random-walk) representation of the threshold dynamics.
    The two panels for $s_n = \pm x$ depict stochastic transition models of the threshold $\theta_n$ viewed as a one-dimensional random walk conditioned on the current signal value.
    In each panel, the transition probabilities of $\theta_n$ depend on the relative position of $\theta_n$ with respect to the dashed boundaries at the points $\pm x$, which determine which arm is selected by the comparison between $s_n$ and $\theta_n$.
    The threshold is updated according to the outcome of the arm selection: $p_A$ ($1-p_A$) and $p_B$ ($1-p_B$) represent that arm $A$ and $B$ win (lose), respectively,
    and the update rules follow Eqs.~\eqref{modeling:eq:updateA} and~\eqref{modeling:eq:updateB}.
    In addition to the threshold transitions, the signal $s_n$ also evolves stochastically;
    the outer arrows indicate the Markov transition of $s_n$, where the sign flips with probability $\gamma$ (see Eq.~\eqref{stochastic:eq:gamma}).    
    }\label{modeling:fig:towsn}
    \end{figure}

Figure~\refsubref{modeling:fig:mabtow}{schematic} illustrates the time-series-based decision making based on the two-valued Markov chain $s_n\in \{-x,\,x\}$.
Here, the $n$-th decision is made according to the inequality between the time series $s_n$ and the threshold $\theta_n$, as discussed in the subsection of the time-series-based decision making.
Specifically, in $s_n = -x$, arm $B$ is selected unless $\theta_n$ is less than $-x$, as shown in the left panel of Fig.~\refsubref{modeling:fig:towsn}{negaposi}.
Conversely, when $s_n$ is $x$, arm $A$ is selected unless $\theta_n$ is greater than $x$, as shown in the right panel of Fig.~\refsubref{modeling:fig:towsn}{negaposi}.
These observations indicate that when $\theta_n$ is less than $-x$, the selected arm is determined to be $A$ in any $s_n$-case.
On the other hand, if $\theta_n$ exceeds $x$, the agent selects arm $B$ regardless of the value of $s_n$.
When the value of $\theta_n$ is between $x$ and $-x$, the selected arm depends on whether $s_n$ is $x$ or $-x$: arm $A$ is selected in the former case, and arm $B$ in the latter.

A primary quantity of interest is the reliability with which the agent selects the optimal arm $A$.  
This is quantified by the \emph{Correct Decision Rate} (CDR), defined as the fraction of times the optimal arm is chosen at the $n$-th decision across independently acting agents.  
In the present stochastic framework, CDR is defined as
\begin{equation}\label{modeling:eq:cdr}
 \mathrm{CDR}_n := \mathbb{P}(a_n = A),
\end{equation}
where $a_n$ is the selected arm at the $n$-th decision.
Arm selection is determined by comparing $s_n$ with the threshold $\theta_n$.  
Hence, it is required to construct a stochastic process model considering variations in both $s_n$ and $\theta_n$.

When $s_n$ is fixed, the evolution of threshold $\theta_n$ can be captured in the framework of one-dimensional random walks with boundaries, as shown in Fig.~\refsubref{modeling:fig:towsn}{rw}.
That is, the position of a walker at time step $n$ corresponds to the threshold $\theta_n$ at the $n$-th decision.
The transition probability is determined by the winning probability associated with the selected arm under the comparison between $s_n$ and $\theta_n$.
When $s_n = -x$, if the threshold $\theta_n$ is greater than $-x$, arm $B$ is selected. 
Then, if arm $B$ wins, which occurs with probability $p_B$, the threshold increases by one as given by Eq.~\eqref{modeling:eq:updateB}, indicating that the walker moves one unit in the positive direction with probability $p_B$.
If not, the threshold decreases by one; that is, the walker makes a unit step in the negative direction with probability $1-p_B$.
If $\theta_n$ is less than $-x$ under $s_n = -x$, the selected arm is $A$, thus the transition probability is determined by the winning probability $p_A$ associated with arm $A$ and the threshold-driven rule given by Eq.~\eqref{modeling:eq:updateA}.
The discussion in the case of $s_n = x$ can be made in the same manner by referring to which arm is selected and how the threshold evolves according to the result of the arm selection.
Here, it is worth noting that the value of the threshold cannot be larger than $N$ nor less than $-N$, as we determined in the previous subsection.
Another important point is that the time series $s_n$ stochastically switches; the sign of $s_n$ changes with probability $\gamma$ for the next decision step.
This indicates that the random walk model also changes with probability $\gamma$, which is critical when the value of the threshold is between $-x$ and $x$, where the transition probability depends on the sign of $s_n$.

To understand the whole system, it is necessary to analyze the joint evolution of $(s_n,\,\theta_n)$.
Indeed, the evolution of $(s_n,\,\theta_n)$ can be described by a Markov process.
The arm selection is only determined by comparing the current signal and threshold, which is not related to the past two values.
Therefore, the threshold dynamics can be formalized independently of the past, as well as signal variation.
Consequently, the overall evolution of the pair $(s_n,\,\theta_n)$ forms a Markov process.
See the \textit{Methods} section for details on this diagram.

Let $\mu_n(\sigma,\,i) := \mathbb{P}((s_n,\,\theta_n) = (\sigma,\,i))$ be the probability that the signal and the threshold are $\sigma\in\{\pm x\}$ and $i\in \{-N,\,\cdots,\,N\}$ on the $n$-th decision, respectively.
Then, the CDR defined in Eq.~\eqref{modeling:eq:cdr} can be represented using $\mu_n(\sigma,\,i)$.
Specifically, CDR is obtained by summing $\mu_n(\sigma,\,i)$ for the pairs $(\sigma,\,i)$ with which the agent chooses arm $A$; that is,
\begin{equation}\label{modeling:eq:cdr2}\begin{split}
	\mathrm{CDR}_n &= \mathbb{P}(a_n = A) = \mathbb{P}(\{(s_n,\,\theta_n)\,|\,s_n \geq \theta_n\})\\
	&= \sum_{i = -N}^{\lfloor x \rfloor} \mu_n(x,\,i) + \sum_{i = -N}^{\lfloor -x \rfloor} \mu_n(-x,\,i).
\end{split}\end{equation}
Here, $\lfloor x \rfloor$ denotes the floor function, i.e., the greatest integer less than or equal to $x$.
The threshold $i$ takes integer values satisfying $-N\le i\le N$, whereas $x$ is assumed to be non-integer; therefore, the conditions $i\le x$ and $i\le -x$ are equivalent to $-N\le i\le \lfloor x\rfloor$ and  $-N\le i\le \lfloor -x\rfloor$, respectively.
These equivalences determine the upper bounds of the two summations in Eq.~\eqref{modeling:eq:cdr2}.
\setcounter{equation}{0}
\section{Results}\label{results}

We evaluate the performance of time-series-based decision making driven by a two-valued Markov chain using the stochastic process model introduced above.  
We present numerical results and discuss the optimal autocorrelation coefficient and the maximized CDR.
We also present a mathematical statement supporting the numerical results in a specific setting and provide additional discussions.

In the following, we assume that the initial threshold is set at the origin, and the initial value of time series is randomly determined:
\begin{equation}
    \mu_{1}(\sigma,\,i) =\begin{cases}
        1/2, & \text{if $(\sigma,\,i) = (\pm x,\,0)$,}\\
        0, & \text{otherwise}.
    \end{cases}
\end{equation}  

\subsection{Numerical results}\label{results:subsec:numerical}
\begin{figure}[ht]
\centering
\begin{minipage}[t]{\textwidth}
	\def\thesubfigure{\Roman{subfigure}}
	\begin{minipage}[t]{0.32\textwidth}
		\includegraphics[width=\textwidth]{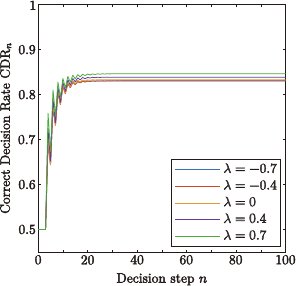}
		\subcaption{$p_B = 0.1$.}\label{results:fig:nVsCDRn_1}
	\end{minipage}\hfill
	\begin{minipage}[t]{0.32\textwidth}
		\includegraphics[width=\textwidth]
        {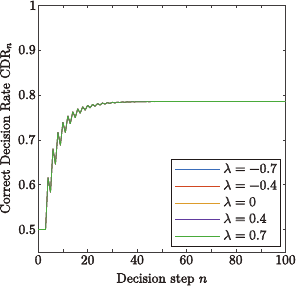}
		\subcaption{$p_B = 0.3$.}\label{results:fig:nVsCDRn_3}
	\end{minipage}\hfill
	\begin{minipage}[t]{0.32\textwidth}
		\includegraphics[width=\textwidth]{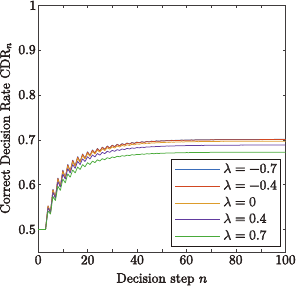}
		\subcaption{$p_B = 0.5$.}\label{results:fig:nVsCDRn_5}
	\end{minipage}
    \caption{Time evolution of the correct decision rate $\mathrm{CDR}_n$ as a function of the decision step $n \in \{1,\,\cdots,\,100\}$ for several autocorrelation coefficients $\lambda$ of the two-valued Markov signal $s_n$. 
    Here, the maximal threshold $N = 4$ and the absolute signal value $x = 3.5$. 
    The winning probability $p_A$ of arm $A$ is fixed to $0.7$, and the winning probability $p_B$ of arm $B$ is set to (I) $0.1$, (II) $0.3$, and (III) $0.5$.
    In all three environments, $\mathrm{CDR}_n$ reaches a near-steady level quickly, so that $\mathrm{CDR}_{1000}$ serves as a proxy for long-term performance.
    It can be seen that the convergent value of $\CDR_n$ is monotonically increasing (decreasing) for the autocorrelation coefficient $\lambda$ of the time series $s_n$ in $p_B = 0.1$ ($0.5$).
    In $p_B = 0.3$, the behavior of $\CDR_n$ is independent of $\lambda$.
    The convergent value of $\CDR_n$ gets smaller when $p_B$ is larger.}\label{results:fig:nVsCDRn}
\end{minipage}\vspace{\baselineskip}\\
\begin{minipage}[t]{\linewidth}
    \centering
    \includegraphics[width=0.4\textwidth]{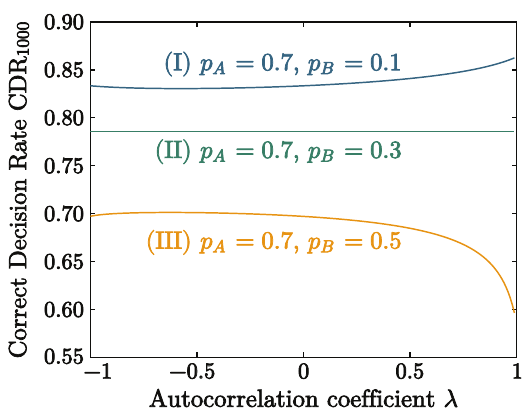}
	\caption{Correct Decision Rate (CDR) at $n = 1000$ over the autocorrelation coefficient $\lambda\in\{-1,\,\cdots,\,0.99\}$ derived by the stochastic process model when $p_A$ is fixed to $0.7$ with $N = 4$ and $x = 3.5$: (\subref{results:fig:nVsCDRn_1}) $p_B = 0.1$, (\subref{results:fig:nVsCDRn_3}) $p_B = 0.3$, (\subref{results:fig:nVsCDRn_5}) $p_B = 0.5$.
	Symbols (I)--(III) correspond to the ones shown in Fig.~\ref{results:fig:nVsCDRn}.
    Negative autocorrelation improves $\mathrm{CDR}_{1000}$ for $p_B=0.5$, whereas positive autocorrelation is favorable for $p_B=0.1$. At the boundary case $p_B=0.3$, $\mathrm{CDR}_{1000}$ is nearly independent of $\lambda$.}\label{results:fig:cdr}
\end{minipage}
\end{figure}

In this subsection, we fix the maximal threshold value $N = 4$ and the absolute signal value $x = 3.5$, respectively.
First, we set $p_A=0.7$. Figures~\refsubref{results:fig:nVsCDRn}{1}--(\subref{results:fig:nVsCDRn_5}) show the relation between the decision step $n$ and correct decision rate $\CDR_n$ for $p_B = 0.1$, $0.3$, $0.5$, respectively.
In each panel, the curves for $\lambda = 0$, $\pm 0.4$, $\pm 0.7$ are compared, 
reminding that $\lambda$ as defined in Eq.~\eqref{modeling:eq:lambda} is the autocorrelation coefficient of the two-valued Markov time series $s_n$.
$\mathrm{CDR}_n$ quickly approaches a steady level within the first several tens of decisions in all three cases.
Comparing these three $p_B$-settings, the $\CDR_n$ decreases when $p_B$ is larger.
This indicates that, when the difference between $p_A$ and $p_B$ is smaller, it is more difficult to determine which arm is more likely to generate a reward.
Moreover, the dependence on $\lambda$ in the long-time value already appears in the early transient.
For $p_B = 0.1$, the convergent value of $\CDR_n$ is monotonically increasing with an increase in $\lambda$, as displayed in Fig.~\refsubref{results:fig:nVsCDRn}{1}.
By contrast, the CDR in $p_B = 0.5$ gets smaller when $\lambda$ is larger, as shown in Fig.~\refsubref{results:fig:nVsCDRn}{5}.
Moreover, for $p_B = 0.3$, all curves overlap, indicating that $\CDR_n$ is independent of $\lambda$.
These facts imply that the positive (negative) autocorrelation coefficient is more effective in the decision-making performance under $p_B = 0.1$ ($0.5$),
and decision-making performance is invariant with respect to the autocorrelation coefficient for $p_B = 0.3$.

Figure~\ref{results:fig:cdr} exhibits the variation of the CDR at the $1000$-th decision step ($\CDR_{1000}$) over the autocorrelation coefficient $\lambda\in \{0,\,\pm 0.01,\,\cdots,\,\pm 0.99,\,-1\}$, which supports the observation in Figs.~\refsubref{results:fig:nVsCDRn}{1}--(\subref{results:fig:nVsCDRn_5})
and serves as a finite-time approximation of the convergent CDR.
The variation of difficulty by changing $p_B$ is also clearly shown; for any $\lambda$, larger winning probability $p_B$ makes $\CDR_{1000}$ smaller.
Moreover a positive autocorrelation coefficient leads to better performance for $p_B = 0.1$.  
By contrast, for $p_B = 0.5$, a negative autocorrelation coefficient improves decision-making performance.  
For $p_B = 0.3$, $\CDR_{1000}$ remains nearly constant as $\lambda$ varies.
These properties are consistent with Figs.~\refsubref{results:fig:nVsCDRn}{1}--(\subref{results:fig:nVsCDRn_5}) for some specific values of $\lambda$, 
and it indicates that the properties are sufficiently generalized for $\lambda$ on the interval $[-1,\ 1)$.

\begin{figure}[ht]
\begin{minipage}[t]{0.41\linewidth}
  \centering
  \includegraphics[width=\linewidth]{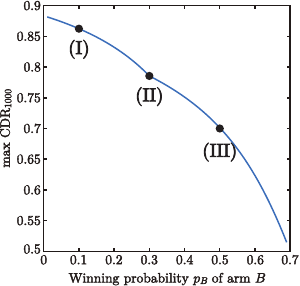}
  \subcaption{Maximal value $\max_{\lambda}\mathrm{CDR}_{1000}$ as a function of $p_B\in\{0.01,\,\cdots,\,0.69\}$ for $p_A=0.7$.}
  \label{results:fig:maxCDR_pB}
\end{minipage}\hfill
\begin{minipage}[t]{0.55\linewidth}
  \centering
  \includegraphics[width=\linewidth]{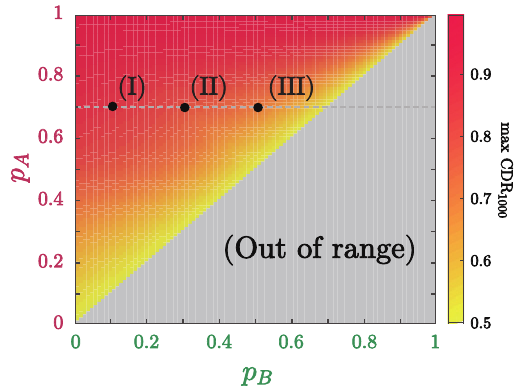}
  \subcaption{Heatmap of the maximum achievable performance  over the environment
  settings $(p_A,\,p_B)\in\{0.01,\,0.02,\,\cdots,\,0.98,\,0.99\}^2$ for $p_A > p_B$.}
  \label{results:fig:maxCDR_heatmap}  
\end{minipage}
\caption{Variation of the maximum achievable performance $\max_{\lambda}\mathrm{CDR}_{1000}$ depending on the environment setting $(p_A,\,p_B)$ with $N=4$ and $x=3.5$.
	Marks (\subref{results:fig:nVsCDRn_1})--(\subref{results:fig:nVsCDRn_5}) correspond to the ones shown in Fig.~\ref{results:fig:nVsCDRn}.
  	Panel (a) shows that, even after optimizing $\lambda$, the best achievable performance decreases as $p_B$ approaches $p_A$, reflecting increasing difficulty in discriminating the better arm.
  Panel (b) shows that the ceiling performance is high when $p_A$ is large and $p_B$ is small. 
  By contrast, it approaches $0.5$ near the diagonal $p_A\approx p_B$, indicating intrinsically hard environments even under optimal $\lambda$.
	}\label{results:fig:maxCDR}
\end{figure}

To clarify how large the best achievable CDR is for a given environment, while keeping $p_A$ fixed, 
we investigate the maximum value of $\CDR_{1000}$ in various $\lambda$, denoted by $\operatorname*{max}_{\lambda}\mathrm{CDR}_{1000}$, as a function of $p_B$ for $p_A=0.7$, which is plotted in Fig.~\refsubref{results:fig:maxCDR}{pB}.
The maximal achievable $\CDR_{1000}$ decreases monotonically as $p_B$ increases toward $p_A$,
indicating that environments with a smaller gap between the winning probabilities are intrinsically more difficult to solve.
Figure~\refsubref{results:fig:maxCDR}{heatmap} shows the corresponding landscape of $\max_{\lambda}\mathrm{CDR}_{1000}$ over the $(p_A,\,p_B)$ plane: the maximum performance is high when $p_A$ is large and $p_B$ is small, and it approaches $0.5$
near the diagonal $p_A= p_B$.

Figures~\refsubref{results:fig:maxCDR}{pB} and (\subref{results:fig:maxCDR_heatmap}) plot $\max_{\lambda}\mathrm{CDR}_{1000}$, i.e., the best achievable performance after optimizing $\lambda$ for each environment. These figures quantify the environment-dependent ceiling of performance.
In particular, the maximal achievable CDR decreases as $p_B$ increases toward $p_A$,
and it approaches $0.5$ near $p_A= p_B$.
This supports the natural interpretation that the discrimination of the optimal arm becomes easier when the winning probabilities are well separated
and becomes harder as the two probabilities approach each other.
At the same time, Figure~\refsubref{results:fig:maxCDR}{heatmap} also indicates that the absolute location of $(p_A,\,p_B)$ matters:
environments near $(p_A,\,p_B)=(0.5,\,0.5)$ are intrinsically ambiguous (maximal CDR close to $0.5$), whereas environments with $p_A$ close to $1$
and $p_B$ close to $0$ are intrinsically easy (maximal CDR close to $1$).

\begin{figure}[ht]
\begin{minipage}[t]{0.41\linewidth}
    \centering
    \includegraphics[width=\textwidth]{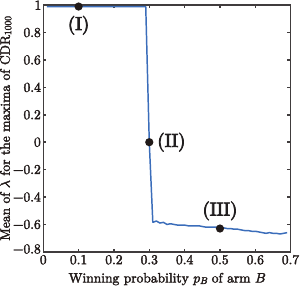}
	\subcaption{Mean of the autocorrelation coefficient $\overline{\lambda}_\mathrm{M}$ that maximizes $\mathrm{CDR}_{1000}$
  as a function of $p_B$ in the range $\lambda\in\{-1,\,\cdots,\,0.99\}$ for $p_A = 0.7$.
}\label{results:fig:lambda_pB}
\end{minipage}\hfill
\begin{minipage}[t]{0.55\linewidth}
    \centering
	\includegraphics[width=\textwidth]{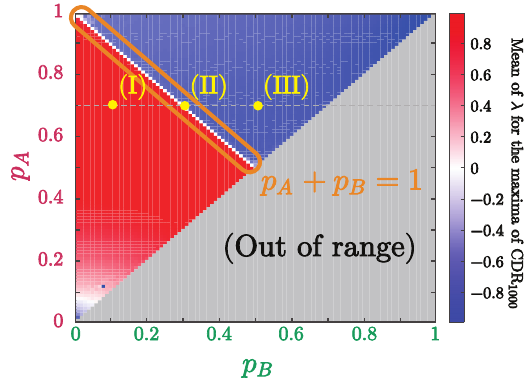}
	\subcaption{Heatmap of the mean of autocorrelation coefficient $\overline{\lambda}_\rM$ for the maxima of CDR at $n = 1000$ for the variation of $(p_A,\,p_B)\in\{0.01,\,0.02,\,\cdots,\,0.98,\,0.99\}^2$ for $p_A > p_B$.
}\label{results:fig:lambda_heatmap}
\end{minipage}
\caption{Variation of the mean of the autocorrelation coefficient $\overline{\lambda}_\rM$ for the maxima of CDR at $n=1000$ depending on the environment setting $(p_A,\,p_B)$ with $N=4$ and $x=3.5$.
	Marks (\subref{results:fig:nVsCDRn_1})--(\subref{results:fig:nVsCDRn_5}) correspond to the ones shown in Fig.~\ref{results:fig:nVsCDRn}.
	Panel (a) shows that the optimal sign switches around $p_B= 0.3$: $\overline{\lambda}_\mathrm{M}>0$ for $p_B<0.3$ and $\overline{\lambda}_\mathrm{M}<0$ for $p_B>0.3$.
	Panel (b) exhibits a phase-like structure with the boundary $p_A+p_B=1$: $\overline{\lambda}_\mathrm{M}<0$ for $p_A+p_B>1$ and $\overline{\lambda}_\mathrm{M}>0$ for $p_A+p_B<1$, while $\overline{\lambda}_\mathrm{M}=0$ on the boundary.
	}\label{results:fig:lambda}
\end{figure}

It is also informative to identify which autocorrelation coefficient $\lambda$ yields the maxima above.
Figure~\refsubref{results:fig:lambda}{pB} plots the mean optimal coefficient $\overline{\lambda}_\mathrm{M}$ as a function of $p_B$
for $p_A=0.7$ with the same $N$ and $x$ as above. The optimal sign of autocorrelation switches sharply around
$p_B = 0.3$, which corresponds to the boundary $p_A+p_B=1$ in this slice:
when $p_A+p_B>1$ (larger $p_B$), the maximum $\CDR_{1000}$ tends to occur for $\overline{\lambda}_\mathrm{M}<0$,
whereas when $p_A+p_B<1$ (smaller $p_B$), the maximum $\CDR_{1000}$ shifts to $\overline{\lambda}_\mathrm{M}>0$.

Figure~\refsubref{results:fig:lambda}{heatmap} further generalizes this observation over the two-dimensional environment space.
It shows a heatmap of the mean autocorrelation coefficient $\overline{\lambda}_\mathrm{M}$ that maximizes $\mathrm{CDR}_{1000}$
for each environment setting $(p_A, p_B)$ with $p_A>p_B$.
It can be observed that when $p_A+p_B>1$ (reward-rich environment), $\mathrm{CDR}_{1000}$ is maximized for $\overline{\lambda}_\mathrm{M}<0$ (blue region in Fig.~\refsubref{results:fig:lambda}{heatmap}).
Conversely, when $p_A+p_B<1$ (reward-poor environment), the maximal $\CDR_{1000}$ occurs for
$\overline{\lambda}_\mathrm{M}>0$ (red region in Fig.~\refsubref{results:fig:lambda}{heatmap}). Moreover, if $p_A+p_B=1$, then $\overline{\lambda}_\mathrm{M}=0$ (white squares in Fig.~\refsubref{results:fig:lambda}{heatmap}), implying that the quality of decision making is unrelated to the autocorrelation coefficient.

A qualitative interpretation of this sign change can be given in terms of exploration--exploitation trade-off:
negative autocorrelation in reward-rich environments tends to induce more frequent sign changes in the signal and thus may promote switching of decisions,
while positive autocorrelation in reward-poor environments tends to stabilize the signal over consecutive steps and may support persistent decisions.
However, this interpretation should be understood as an intuition rather than a mathematical explanation, because $\lambda$ is a property of the signal statistics and not an explicit control parameter of an exploration policy in the present formulation.

\subsection{Mathematical statement}\label{results:subsec:mathematical}

The following theorem rigorously characterizes the special boundary case $p_A+p_B=1$, which underlies the $\lambda$-independent behavior
observed in Figs.~\refsubref{results:fig:nVsCDRn}{3} and \ref{results:fig:cdr}, and the appearance of $\overline{\lambda}_\mathrm{M}=0$ in
Figs.~\refsubref{results:fig:lambda}{pB} and (\subref{results:fig:lambda_heatmap}).

\begin{thm}\label{results:thm}
In the time-series-based decision making with the two-valued time series $s_n \in \{\pm x\}$, whose autocorrelation coefficient is $\lambda \in [-1,\,1)$, the correct decision rate $\CDR_n$ under the condition $p_A + p_B = 1$ converges as $n \to \infty$ to
\begin{equation}\label{results:eq:CDRinfty}
	\CDR_\infty := \lim_{n\to\infty} \CDR_n = \fraction{2p_A^{2N+1} -p_A^{N-\lfloor x \rfloor}(1-p_A)^{N+\lfloor x \rfloor+1} -p_A^{N+\lfloor x \rfloor+1}(1-p_A)^{N-\lfloor x \rfloor}}{2\{p_A^{2N+1} -(1-p_A)^{2N+1}\}}.
\end{equation}
\end{thm}
Note that this result for $\CDR_\infty$ is clearly theoretical; this is in contrast to $\CDR_{1000}$ in the earlier section, which is obtained in numerical calculations.
For example, if $p_A = 0.7$ and thus $p_B = 0.3$, the limiting value of $\CDR$ is $\CDR_\infty\approx 0.7855$.
This is consistent with the numerical result in Fig.~\ref{results:fig:cdr}, where $\CDR_{1000}$ is numerically obtained as a finite-time approximation of $\CDR_\infty$.
This theorem indicates that the decision-making performance is completely independent of the autocorrelation coefficient $\lambda$ when $p_A + p_B = 1$, which coincides with the observation that $\overline{\lambda}_{\rM} = 0$ in Fig.~\refsubref{results:fig:lambda}{heatmap}.
The proof of Theorem~\ref{results:thm} is found in the \textit{Methods} section.

Further analyses of the case $p_A + p_B = 1$ produce interesting properties from the perspective of the ``difficulty'' of decision making.
For example, the convergent values of $\CDR_\infty$, given in Eq.~\eqref{results:eq:CDRinfty}, satisfy
\begin{equation}
	\lim_{p_A\to 0.5} \CDR_\infty = \fraction{1}{2},
	\qquad
	\lim_{p_A\to 1} \CDR_\infty = 1.    
\end{equation}
Moreover, by denoting the difference between $N$ and $\lfloor x \rfloor$ by a natural number $m = N - \lfloor x \rfloor$ and taking the limit $N\to\infty$ for $0.5 < p_A < 1$, we obtain the convergent formula
\begin{equation}
	\lim_{N\to \infty}\CDR_\infty = 1 -\fraction{1}{2}\left(\fraction{1-p_A}{p_A}\right)^m = 1 -\fraction{1}{2}\left(\fraction{p_B}{p_A}\right)^m.
\end{equation}	
This indicates that if $N$ is sufficiently large, $\CDR_\infty$ as a function of $p = p_A$ can be approximated by a function $f_m(p)$ ($0.5<p<1$) defined by
\begin{equation}\label{results:eq:fp}
    f_m(p) = 1-\fraction{1}{2}\left(\fraction{1-p}{p}\right)^m.
\end{equation}

\begin{figure}[ht]
\begin{minipage}[t]{0.48\linewidth}
    \centering
    \includegraphics[width=\textwidth]{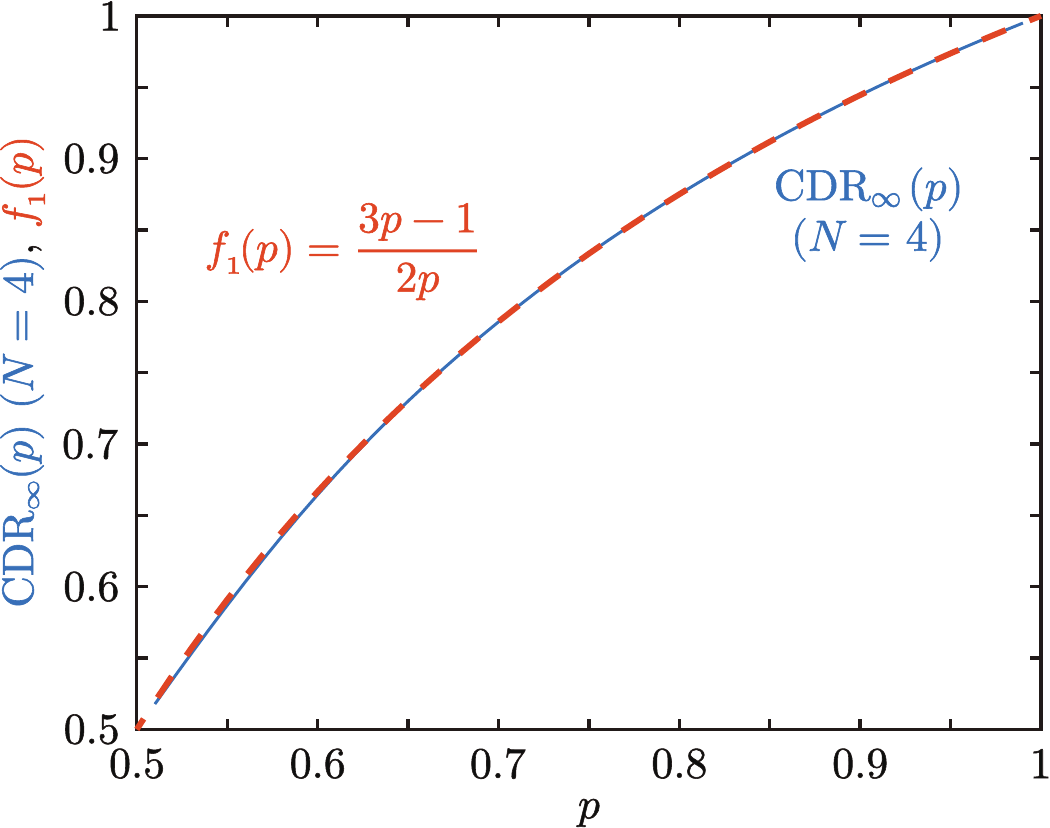}
	\subcaption{
	$(N,\,x) = (4,\,3.5)$, leading to $m=1$.
	}\label{results:fig:fp_m1}
\end{minipage}
\hfill
\begin{minipage}[t]{0.48\linewidth}
    \centering
    \includegraphics[width=\textwidth]{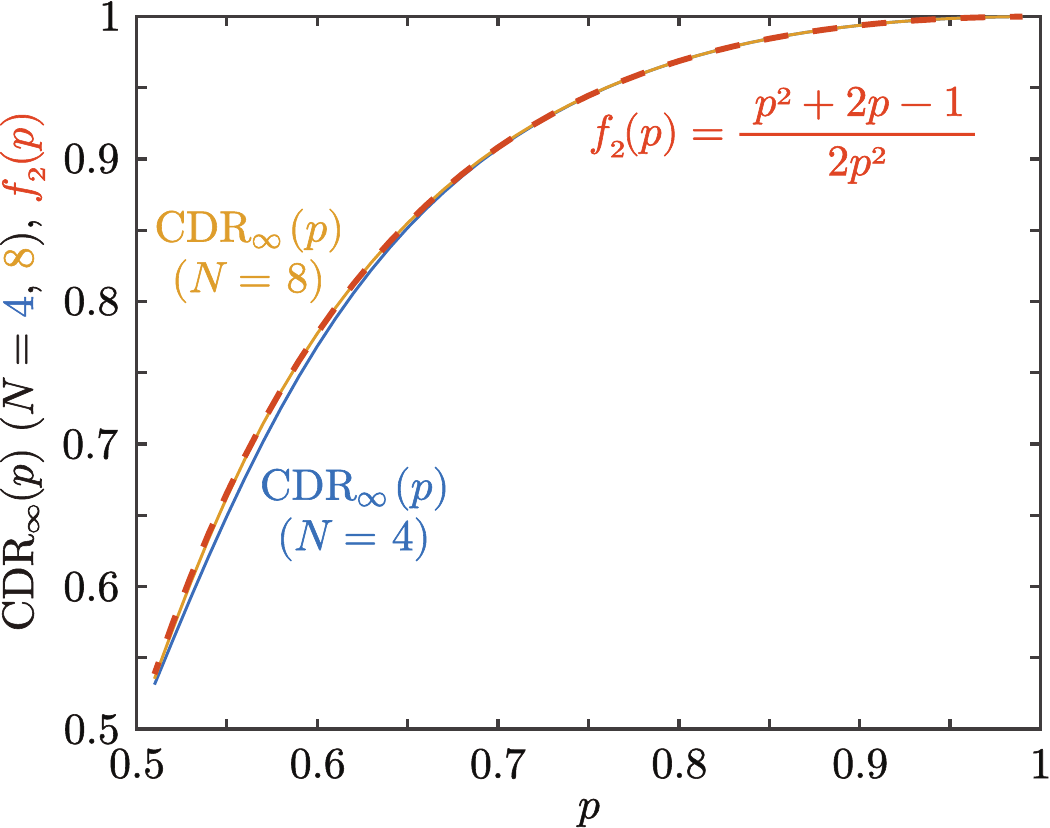}
	\subcaption{
	$(N,\,x) = (4,\,2.5),\,(8,\,6.5)$, both leading to $m=2$.
	}\label{results:fig:fp_m2}
\end{minipage}
\caption{
	Curves of $\CDR_\infty$ as a function of $p = p_A \in (0.5,\,1)$ (see Eq.~\eqref{results:eq:CDRinfty}) in comparison with $f_m(p) = 1-(1-p)^m/(2p^m)$ (see Eq.~\eqref{results:eq:fp}) with $m = N -\lfloor x \rfloor$.
	It can be seen that $f_m(p)$ approximates $\CDR_\infty$ well, 
	and panel (b) demonstrates that $\mathrm{CDR}_\infty$ approaches $f_m(p)$ as $N$ increases.
	}\label{results:fig:fp}
\end{figure}

Figures~\refsubref{results:fig:fp}{m1} and (\subref{results:fig:fp_m2}) show the variations of $\CDR_\infty(p)$ for $p\in (0.5,\,1)$ in representative cases, together with the approximating functions $f_m(p)$ for $m=1$ and $2$, respectively.
The case of $m=1$ demonstrates that $f_m(p)$ approximates $\CDR_\infty$ well, as shown in Fig.~\refsubref{results:fig:fp}{m1}.
The case of $m=2$ also exhibits such an approximation, as well as the clearer agreement for larger $N$ by comparing $N=4$ and $N=8$ cases, as shown in Fig.~\refsubref{results:fig:fp}{m2}.
These results represent the relationship between the environment setting and the difficulty of decision making.
First, when $p_A$ is close to $0.5$, which implies that $p_B$ is also close to $0.5$, the difference in winning probabilities between arms $A$ and $B$ becomes small,
making it difficult to determine which arm yields rewards more frequently.
As a result, the agent inevitably selects an arm almost at random,
which corresponds to a probability of $0.5$ of a correct decision.
By contrast, if $p_A$ is close to $1$, and consequently $p_B$ is close to $0$, 
the difference in winning probabilities between arms $A$ and $B$ is large,
meaning that it is easy to identify arm $A$ as the better one.
As a result, the agent eventually always selects arm $A$.
Moreover, $\CDR_\infty$ with a sufficiently large threshold value $N$ is monotonically increasing for $p_A \in (0.5,\,1)$ since $p_B/p_A < 1$.
This corresponds to the fact that for a larger $p_A$, it is easier for the agent to identify which arm has the larger winning probability.

\setcounter{equation}{0}
\section{Discussion}\label{discussion}

We have presented the performance of time-series-based decision making via the two-valued Markov chain by calculating the associated stochastic process model,
clarifying that the optimal autocorrelation coefficient depends on how the winning probabilities of arms $A$ and $B$ are set.
It should be recalled that the previous work~\cite{okada2022theory} claimed that the performance of the time-series-based decision making improves if the autocorrelation coefficient inherent in the time series is negative.
However, our results clarified that the negativity of the autocorrelation coefficient does not always enhance the decision-making performance in this model.
The prior research~\cite{okada2022theory} reported that this benefit is especially noticeable when the winning probabilities of the two arms $A$ and $B$ are $(p_A, p_B) = (0.9, 0.3)$, $(0.6, 0.5)$, or $(0.9, 0.7)$, with $N$ equal to 2 or 4 and $x$ satisfying $N-1 < x < N$, by calculating the CDR at decision step $n=1000$ as a function of the autocorrelation coefficient $\lambda$.  
Specifically, it was reported that the case $(p_A,\,p_B) = (0.9,\,0.3)$ exhibits the highest CDR at $\lambda = -0.5$, while the cases $(p_A,\,p_B) = (0.6,\,0.5)$ and $(0.9,\,0.7)$ reach their maxima at $\lambda = -0.6$.
However, all of these cases satisfy $p_A + p_B > 1$, meaning that the previous work~\cite{okada2022theory} did not examine the cases $p_A + p_B \leq 1$, even though the sum of $p_A$ and $p_B$ fundamentally affects whether negative or positive autocorrelation improves performance.

It should be remarked that this model is not completely identical to the model introduced in the experimental reports using laser-chaos-based decision making~\cite{naruse2017ultrafast}.
In that report, a parameter $\Omega$, applied for increment or decrement when the arm loses (see Eqs.~\eqref{methods:eq:TAA} and \eqref{methods:eq:TAB} in the \textit{Methods} section), is defined using the sum of the winning probabilities of arms $A$ and $B$; applying our notations,
\begin{equation}\label{discussion:eq:Omega}
	\Omega = \fraction{p_A + p_B}{2 - (p_A +p_B)}.
\end{equation}
In the present model, $\Omega$ is fixed at $1$ 
 so that we can make discussions only for integer threshold values.
The difference between the model and the experimental results appears even in the case of $p_A + p_B = 1$, which leads to $\Omega = 1$ in Eq.~\eqref{discussion:eq:Omega}, coinciding with our setting.
Our results in this case indicate that enhancing the decision-making performance is completely independent of the autocorrelation coefficient of the time series, which differs from the results shown with laser-chaos-based decision making~\cite{naruse2017ultrafast}.
Potential reasons for this difference include that the time series applied in the stochastic process model is simplified compared with those generated from chaotic laser dynamics.
In particular, the prior studies~\cite{naruse2017ultrafast, okada2021analysis} demonstrate that the histogram of signal values in chaotic time series shows a Gaussian-like shape with slight skewness between the positive and negative sides, whereas that of the two-valued Markov chain converges to a scaled Rademacher distribution $\mathbb{P}(s_n = \pm x) \approx 1/2$ for sufficiently large $n$ and for any initial distribution $\mathbb{P}(s_1 = \pm x)$.
This indicates that the time series in the mathematical model analyzed in this paper has different statistical properties,
which may result in different autocorrelation characteristics.

Comparing the theoretical CDR in time-series-based decision making with those empirically obtained by representative classical bandit algorithms also provides a useful perspective on the present stochastic-process model.
When classical policies such as softmax~\cite{sutton2018reinforcement, daw2006cortical} and upper confidence bound~\cite{auer2002finite} are appropriately tuned, they can achieve higher long-time CDR than the current time-series-based decision making described by the simplified stochastic-process model.
This indicates that the present stochastic-process model has room for improvement in long-time decision accuracy.
However, the stochastic-process model exhibits rapid early-stage growth of CDR as we see in Fig.~\ref{results:fig:nVsCDRn}, suggesting that it captures an adaptive-response aspect. 
This point is worth emphasizing because it is expected that the stochastic process model describing time-series-based decision making formalizes the process of laser-chaos-based decision makers, 
and fast adaptive decision making has been demonstrated using ultrafast chaotic time series in the laser-chaos-based decision making \cite{naruse2017ultrafast}. 

This comparison gives a constructive perspective on future extensions of the stochastic process framework.
It suggests that the present model should be extended so as to retain the rapid early-stage adaptation captured by the current formulation while improving long-time decision accuracy.
In this sense, the comparison further supports the importance of extending the present model toward more realistic and practical forms of time-series-based decision making.

Our results indicate that the relation between the autocorrelation coefficient of a time series and decision making can depend on the environment setting (i.e., winning probabilities).
Thus, it is necessary to pursue what kind of time series with a negative autocorrelation enhances decision-making performance.
One strategy is to conduct successive research starting from the two-valued Markov chain case while relaxing the simplifying assumptions introduced in the present model.
For example, how does the decision-making performance change if we reduce the quantity of memory?
The threshold adjuster in the laser-chaos-based decision maker can include a memory parameter $\alpha$, which appears in Eqs.~\eqref{methods:eq:TAA} and \eqref{methods:eq:TAB} in the \textit{Methods} section.

As mentioned in the \textit{Methods} section, our model assumes $\alpha = 1$, which indicates that the most recent memory is completely conserved.
This assumption is another factor that makes the present model different from the practical laser-chaos-based decision maker.
The memory parameter is crucial to be tuned for providing opportunities for exploration.
Indeed, the prior investigation has clarified that the optimal memory parameter is near one but not just one for several cases, which seems to be claimed in general cases \cite{mihana2018memory}.
Thus, it is important to analyze decision making utilizing a threshold value whose evolution formula includes the memory parameter.
It is expected that a corresponding stochastic process model replicating such decision making will be based on a random walk whose position is pulled back at each time step;
such random walks have been used as the antlion random walk~\cite{narimatsu2025study}, Bernoulli convolution~\cite{solomyak2004notes}, and the autoregressive sequence of order~1 (AR(1))~\cite{hinrichs2020persistence}.

Another direction for extending the current study is to investigate the dependency of decision-making performance on the lag of the autocorrelation coefficient.
Currently, we consider only signals with autocorrelation coefficient of lag one.
A longer lag could be useful when obtaining the autocorrelation coefficient.
The relationship between the sign of the autocorrelation coefficient and decision-making performance may change in the presence of different lag.

It is necessary to consider extending the current study to time-series-based decision making for bandit problems with three or more arms to make our contribution more practical.
It has been reported that the laser-chaos-based decision maker can be applied to settings with three or more arms \cite{naruse2018scalable}.
In this report, scalability for the number of arms is realized by generating a $K$-bit binary series from the chaotic laser dynamics and choosing one of the $2^K$ arms according to the binary series.
This $K$-bit binary series is generated through multiple comparisons between the threshold and chaotic time series; the two-armed case is a special case of this scheme ($K=1$).
However, the full analytical treatment of the multi-threshold, multi-bit stochastic process remains an open problem,
and we will work on it in the near future.

\setcounter{equation}{0}
\section{Conclusion}\label{conclusion}
In this paper, we analyzed a stochastic process model to elaborate the decision-making process in the two-armed bandit problem utilizing a stochastic time series based on the tug-of-war principle.
We investigated the relation between the autocorrelation coefficient and the decision-making performance.
The stochastic process model was constructed by a Markov chain describing the variation of the time series and the adjustable threshold value,
where it was assumed that the time series was a two-valued Markov chain for simplicity.
The numerical results clarified that the relation between the autocorrelation coefficient and decision-making performance depends on the environment setting:
if the sum of the winning probabilities associated with the two arms is larger than one (reward-rich environment), the negative autocorrelation coefficient improves the decision-making performance.
On the other hand, when the sum is less than one (reward-poor environment), a positive autocorrelation coefficient is more effective in enhancing the performance.
Moreover, when the sum exactly equals one, the decision-making performance becomes independent of the autocorrelation coefficient of the time series.
This observation was also demonstrated by a mathematical statement.

Future research should focus on understanding which specific properties of real time series contribute to performance enhancement, including autocorrelation, memory, and lag effects.
Our findings would pave the way for understanding the mechanism of performance improvement in time-series-based decision making for solving the two-armed bandit problem and for future extensions to multi-armed settings.

\setcounter{equation}{0}
\section{Methods}\label{methods}

\subsection{Relation between the time-series-based decision making and the laser-chaos-based decision making}\label{methods:subsec:laser}

This subsection explains how the time-series-based decision making framework presented in the \textit{Modeling} section corresponds to the laser-chaos-based decision maker \cite{naruse2017ultrafast}.  
In that system, the time series $(s_n)$ is obtained by sampling the chaotic signal generated by a semiconductor laser with optical feedback.  
Although the underlying dynamics is continuous, discrete sampling produces the sequence used for decision making.

The threshold $\theta_n$ is determined by a variable $\mathrm{TA}_n$, referred to as the \emph{threshold adjuster}.  
The relation between $\theta_n$ and $\mathrm{TA}_n$ is given by
\begin{equation}
	\theta_n = k\,[\mathrm{TA}_n],
\end{equation}
where $k \in \mathbb{R}$ is a constant.

The update rule for $\mathrm{TA}_n$ in the laser-chaos-based decision maker is defined as follows.  
Under selection of arm $A$,
\begin{equation}\label{methods:eq:TAA}
	\mathrm{TA}_{n+1} =
	\begin{cases}
		\max\{ \alpha\,\mathrm{TA}_n - \Delta,\,-N\} & \text{if arm $A$ wins},\\
		\min\{ \alpha\,\mathrm{TA}_n + \Omega,\, N\} & \text{if arm $A$ loses},
	\end{cases}
\end{equation}
and under selection of arm $B$,
\begin{equation}\label{methods:eq:TAB}
	\mathrm{TA}_{n+1} =
	\begin{cases}
		\min\{ \alpha\,\mathrm{TA}_n + \Delta,\, N\} & \text{if arm $B$ wins},\\
		\max\{ \alpha\,\mathrm{TA}_n - \Omega,\,-N\} & \text{if arm $B$ loses}.
	\end{cases}
\end{equation}
Here, $\alpha \in [0,1]$ is a memory parameter.  
The threshold adjuster is increased or decreased by $\Delta$ when the selected arm wins, and by $\Omega$ when it loses.

By setting $k=\alpha=\Delta=\Omega=1$ and the initial threshold adjuster $\mathrm{TA}_1=0$, one obtains $\theta_n = \mathrm{TA}_n$ for every decision step $n$.  
In this case, the laser-chaos-based decision making coincides exactly with the time-series-based decision making described in this paper.

\subsection{Details on constructing the stochastic-process model}\label{methods:subsec:details}

This subsection gives detailed discussions on the state transition of the pair $(s_n,\,\theta_n)$ of the signal and the threshold introduced in the \textit{Modeling} section.

The one-step probability of $(s_n,\,\theta_n)$ is transformed as follows:
\begin{equation}\label{methods:eq:transition}\begin{split}
	\mathbb{P}((s_{n+1},\,\theta_{n+1}) = (\tau,\,j)\,|\,(s_n,\,\theta_n) = (\sigma,\,i)) \hspace{-15em}\vrule width0pt depth7pt&\\
		&= \fraction{\mathbb{P}((s_{n+1},\,\theta_{n+1},\,s_n,\,\theta_n) = (\tau,\,j,\,\sigma,\,i))}{\mathbb{P}((s_n,\,\theta_n) = (\sigma,\,i))}\\
		&= \fraction{\mathbb{P}(s_{n+1} = \tau\,|\,(\theta_{n+1},\,s_n,\,\theta_n) = (j,\,\sigma,\,i))\ \mathbb{P}((\theta_{n+1},\,s_n,\,\theta_n) = (j,\,\sigma,\,i))}{\mathbb{P}((s_n,\,\theta_n) = (\sigma,\,i))}\\
		&= \mathbb{P}(s_{n+1} = \tau\,|\,(\theta_{n+1},\,s_n,\,\theta_n) = (j,\,\sigma,\,i))\ \mathbb{P}(\theta_{n+1} = j\,|\,(s_n,\,\theta_n) = (\sigma,\,i))\\
		&= \mathbb{P}(s_{n+1} = \tau\,|\,s_n = \sigma)\ \mathbb{P}(\theta_{n+1} = j\,|\,(s_n,\,\theta_n) = (\sigma,\,i)).
\end{split}\end{equation}
Here, the variation of the signal is independent of the threshold, leading to 
\begin{equation}
    \mathbb{P}(s_{n+1} = \tau\,|\,(\theta_{n+1},\,s_n,\,\theta_n) = (j,\,\sigma,\,i)) = \mathbb{P}(s_{n+1} = \tau\,|\,s_n = \sigma),
\end{equation}
which is applied in the transformation above. 
The time series $(s_n)$ is a two-valued Markov chain given by Eq.~\eqref{stochastic:eq:gamma}, 
and $\mathbb{P}(s_{n+1} = \tau\,|\,s_n = \sigma)$ is described using the signal-changing probability $\gamma$ as
\begin{equation}\label{methods:eq:s}
	 \mathbb{P}(s_{n+1} = \tau\,|\,s_n = \sigma) 
	 	= \begin{cases} 1-\gamma, & \tau = \sigma,\\
			\gamma, & \tau = -\sigma.
			\end{cases}
\end{equation}

Next, we consider the probability $\mathbb{P}(\theta_{n+1} = j\,|\,(s_n,\,\theta_n) = (\sigma,\,i))$.
The variation of the threshold is governed by which arm is selected and whether the arm wins or loses;
the former is fixed by comparing which $s_n$ or $\theta_n$ is larger,
and the latter is determined by the probabilistic outcome associated with the winning probability of the selected arm, see Eqs.~\eqref{modeling:eq:updateA} and \eqref{modeling:eq:updateB}.
In any case, the threshold changes by $\pm 1$ unless it is at the boundaries $\pm N$, in which case it remains.  
That is, there exists a map $q:\{\pm x\}\times\{-N,\,\cdots,\,N\}\to[0,\,1]$ satisfying
\begin{equation}\label{methods:eq:threshold}
 \mathbb{P}(\theta_{n+1}=j\mid(s_n,\,\theta_n)=(\sigma,\,i))
  =\begin{cases}
    q(\sigma,\,i), & j=\min\{i+1,\,N\},\\[2pt]
    1-q(\sigma,\,i), & j=\max\{i-1,\,-N\},\\[2pt]
    0, & \text{otherwise}.
   \end{cases}
\end{equation}
Here, $q(\sigma,\,i)$ is interpreted as the probability that $\theta_n$ moves upward from $i$ under signal $\sigma$.  
From the selection rule in {\bsf[STEP 1]} in the \textit{Modeling} section,
\begin{equation}\label{methods:eq:q}
 q(\sigma,\,i)=
  \begin{cases}
   1-p_A,& \sigma> i,\\[2pt]
   p_B,& \sigma< i.
  \end{cases}
\end{equation}

\begin{figure}[htp]
	\centering
	\includegraphics[width=0.8\textwidth]{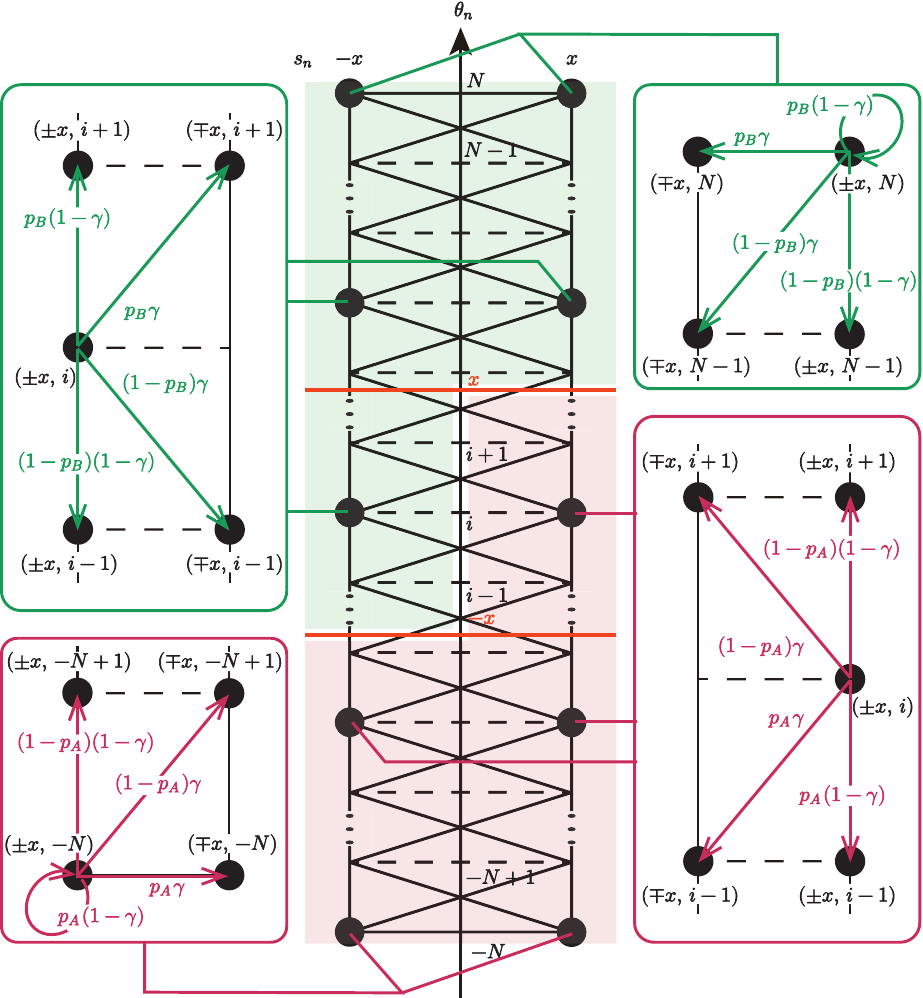}
	\caption{
    State-space transition diagram of the coupled Markov process $(s_n,\,\theta_n)$ describing the time-series-based decision making.
    Each node represents a joint state $(s_n,\,\theta_n)=(\sigma,\,i)$, where the signal takes one of two values $\sigma\in\{\pm x\}$, and the threshold $i\in\{-N,\,\cdots,\,N\}$ is bounded.
    The colored regions indicate which arm is selected by the comparison between $\sigma$ and $i$: 
    nodes in the red region satisfy $\sigma > i$ (arm $A$ is chosen), whereas nodes in the green region satisfy $\sigma < i$ (arm $B$ is chosen).
    From any node, the next step is determined by two stochastic components: the signal either stays the same or flips sign with probabilities $1-\gamma$ and $\gamma$, and the threshold moves to $i\pm 1$ with winning/losing probability $(p_A,\,1-p_A)$ and $(p_B,\,1-p_B)$ according to the arm selected by comparing $\sigma$ and $i$ and its reward result.
    The arrows explicitly enumerate all one-step transitions $(\sigma,\,i)\to(\tau,\,j)$, i.e., signal changes $\sigma\to\tau$ combined with threshold updates $i\to j$, including the boundary cases at $i=\pm N$.
    	}\label{modeling:fig:diagram}
\end{figure}

It is worth noting that a two-dimensional state-space representation, as shown in Fig.~\ref{modeling:fig:diagram}, helps clarify the transition structure of the coupled process $(s_n,\,\theta_n)\in \{\pm x\}\times \{-N,\,-N+1,\,\cdots,\,N-1,\,N\}$.
In this diagram, horizontal moves correspond to changes in the signal $s_n$, whereas vertical moves correspond to updates of the threshold $\theta_n$.
As a concrete example, we consider the state $(s_n,\,\theta_n) = (x,\,i)$ with $-x<i<x$.
In this case, the agent selects arm $A$ since the signal exceeds the threshold.
Then, four transitions to the next state are possible, with probabilities
\begin{equation}
    \mathbb{P}((s_{n+1},\,\theta_{n+1}) = (\tau,\,j)\,|\,(s_n,\,\theta_n) = (x,\,i)) = 
    \begin{cases}
        (1-p_A)(1-\gamma), & (\tau,\,j) = (x,\,i+1),\\
        p_A(1-\gamma), & (\tau,\,j) = (x,\,i-1),\\
        (1-p_A)\gamma, & (\tau,\,j) = (-x,\,i+1),\\
        p_A\gamma, & (\tau,\,j) = (-x,\,i-1).\\
    \end{cases}
\end{equation}
For $s_{n+1} = x$, where the signal does not flip, only a vertical move occurs: the threshold increases to $i+1$ when arm $A$ loses and decreases to $i-1$ when it wins, corresponding to the upward and downward arrows in the lower-right panel of Fig.~\ref{modeling:fig:diagram}.
In contrast, for $s_{n+1} = -x$, where the signal flips, the transition involves both a horizontal in $s_n$ and a vertical update of $\theta_n$, producing diagonally directed arrows in the same panel.

\subsection{Recurrence relation on the 
state transition}\label{methods:subsec:details}

This subsection discusses the recurrence relation of the probability $\mu_n(\sigma,\,i)$ on the state $(s_n,\,\theta_n) = (\sigma,\,i)$ and obtain its matrix--vector expression.
The evolution of $\mu_n$ is governed by
\begin{equation}\label{methods:eq:mu}\begin{split}
	\mu_{n+1}(\sigma,\,i) &= \mathbb{P}((s_{n+1},\,\theta_{n+1}) =(\sigma,\,i))\\
		&= \sum_{\tau\in\{\pm x\}}\sum_{j = -N}^{N} \mathbb{P}((s_{n+1},\,\theta_{n+1}) = (\sigma,\,i)\,|\,(s_n,\,\theta_n) = (\tau,\,j))\ \mathbb{P}((s_n,\,\theta_n) = (\tau,\,j))\\
		&= \sum_{\tau\in\{\pm x\}}\sum_{j=-N}^{N} \mathbb{P}(s_{n+1} = \sigma\,|\,s_n = \tau)\  \mathbb{P}(\theta_{n+1} = i\,|\,(s_n,\,\theta_n) = (\tau,\,j))\ \mu_n(\tau,\,j),
\end{split}\end{equation}
wherein the third transform holds by Eq.~\eqref{methods:eq:transition}.

First, assume that $i\not = -N\,\text{or}\,N$; by substituting Eqs.~\eqref{methods:eq:s}--\eqref{methods:eq:q} to Eq.~\eqref{methods:eq:mu} with $\sigma = \pm x$, we have
\begin{equation}\begin{split}
	\mu_{n+1}(\pm x,\,i) &= \sum_{\tau\in\{\pm x\}}\sum_{j=-N}^{N} \mathbb{P}(s_{n+1} = \pm x\,|\,s_n = \tau)\ \mathbb{P}(\theta_{n+1} = i\,|\,(s_n,\,\theta_n) = (\tau,\,j))\ \mu_n(\tau,\,j)\\
		&= (1-\gamma)q(\pm x,\,i-1)\mu_n(\pm x,\,i-1) +\gamma q(\mp x,\,i-1)\mu_n(\mp x,\,i-1)\\
		&\hspace{2em} +(1-\gamma)(1-q(\pm x,\,i+1))\mu_n(\pm x,\,i+1) +\gamma (1-q(\mp x,\,i+1))\mu_n(\mp x,\,i+1).
\end{split}\end{equation}
Here, we introduce the notations
\begin{align}
 \bfit{\mu}_n(i) &= \left[\begin{array}{l} \mu_n(x,\,i)\\[1pt] \mu_n(-x,\,i) \end{array}\right],\\[2pt]
 \Gamma &= \begin{bmatrix} 1-\gamma & \gamma\\[1pt] \gamma & 1-\gamma \end{bmatrix},\\
 Q(i)&=\begin{bmatrix} q(x,\,i) & 0\\[1pt] 0 & q(-x,\,i) \end{bmatrix}, \label{methods:eq:Q}\\
 P(i) &=\begin{bmatrix} 1-q(x,\,i) & 0\\[1pt] 0 & 1-q(-x,\,i) \end{bmatrix}. \label{methods:eq:P}
\end{align}
Then, the discussions on $\mu_n(x,\,i)$ and $\mu_n(-x,\,i)$ can be summarized as those of $\bfit{\mu}_n(i)$ as follows:
\begin{equation}\label{methods:eq:mu:nedge}\begin{split}
	\bfit{\mu}_{n+1}(i) &= \twobytwo{(1-\gamma)q(x,\,i-1)}{\gamma q(-x,\,i-1)}{\gamma q(x,\,i-1)}{(1-\gamma) q(-x,\,i-1)}\bfit{\mu}_n(i-1)\\
	&\hspace{8em} + \twobytwo{(1-\gamma)(1-q(x,\,i+1))}{\gamma (1-q(-x,\,i+1))}{\gamma (1-q(x,\,i+1))}{(1-\gamma) (1-q(-x,\,i+1))}\bfit{\mu}_n(i+1)\\
	&= \Gamma (Q(i-1)\bfit{\mu}_n(i-1) + P(i+1)\bfit{\mu}_n(i+1)).
\end{split}\end{equation}
At the lower boundary $i=-N$, the threshold may stay: 
\begin{equation}\begin{split}
	\mu_{n+1}(\pm x,\, -N) &= (1-\gamma)(1-q(\pm x,\,-N))\mu_n(\pm x,\,-N) +\gamma (1-q(\mp x,\,-N))\mu_n(\mp x,\,-N)\\
		&\phantom{=} +(1-\gamma)(1-q(\pm x,\,-N+1))\mu_n(\pm x,\,-N+1) +\gamma (1-q(\mp x,\,-N+1))\mu_n(\mp x,\,-N+1),
\end{split}\end{equation}
and its corresponding matrix--vector expression is
\begin{equation}\label{methods:eq:mu:middle}\begin{split}
	\bfit{\mu}_{n+1}(-N) &= \twobytwo{(1-\gamma)(1-q(x,\,-N))}{\gamma(1-q(-x,\,-N))}{\gamma(1-q(-x,\,-N))}{(1-\gamma) (1-q(x,\,-N))}\bfit{\mu}_n(-N)\\
	&\hspace{2em} + \twobytwo{(1-\gamma)(1-q(x,\,-N+1))}{\gamma (1-q(-x,\,-N+1))}{\gamma (1-q(x,\,-N+1))}{(1-\gamma) (1-q(-x,\,i+1))}\bfit{\mu}_n(-N+1)\\
	&= \Gamma (P(-N)\bfit{\mu}_n(-N) + P(-N+1)\bfit{\mu}_n(-N+1))\\
	&= \Gamma (p_A \bfit{\mu}_n(-N) +P(-N+1)\bfit{\mu}_n(-N+1)).
\end{split}\end{equation}
It should be noted that the staying transition probability on the threshold value is processed by the second condition in Eq.~\eqref{methods:eq:threshold}.
The second transform comes from the above-mentioned fact that $q(\sigma,\,-N) = 1 -p_A$ for both cases of $\sigma = \pm x$, which leads to $P(-N) = p_A I_2$ with the identity matrix $I_d$ of order $d$.

Similarly, we can consider the staying on the threshold value $N$ in discussing $\mu_n(\pm x,\,N)$.
Specifically, $\mu_n(\pm x,\,N)$ is expanded as 
\begin{equation}\begin{split}
	\mu_{n+1}(\pm x,\, N) &= (1-\gamma)q(\pm x,\,N-1)\mu_n(\pm x,\,N-1) +\gamma q(\mp x,\,N-1)\mu_n(\mp x,\,N-1)\\
		&\hspace{8em} +(1-\gamma)q(\pm x,\,N)\mu_n(\pm x,\,N) +\gamma q(\mp x,\,N)\mu_n(\mp x,\,N),
\end{split}\end{equation}
and its matrix--vector expression is
\begin{equation}\label{methods:eq:mu:pedge}\begin{split}
	\bfit{\mu}_{n+1}(N) &= \twobytwo{(1-\gamma)q(x,\,N-1)}{\gamma q(-x,\,N-1)}{\gamma q(x,\,N-1)}{(1-\gamma)q(-x,\,N-1)}\bfit{\mu}_n(N-1)\\
	&\hspace{8em} + \twobytwo{(1-\gamma)q(x,\,N)}{\gamma q(-x,\,N)}{\gamma q(x,\,N)}{(1-\gamma) q(-x,\,N+1)}\bfit{\mu}_n(N)\\
	& = \Gamma (Q(N-1)\bfit{\mu}_n(N-1) +Q(N)\bfit{\mu}_n(N))\\
	& = \Gamma (Q(N-1)\bfit{\mu}_n(N-1) +p_B\bfit{\mu}_n(N)).
\end{split}\end{equation}

\subsection{Proof of Theorem~\ref{results:thm}}\label{methods:subsec:proof}

First, we consider the transition probability matrix $M$ for the state transition diagram describing the time-series-based decision making driven by a two-valued Markov chain.  
Let $\bfit{\mu}_n$ be the vector representing the probability distribution $\mu_n(\sigma,\,i)$ for the pair composed of the signal $\sigma$ and the threshold $i$:
\begin{equation}
	\bfit{\mu}_n := \left[\begin{array}{l} \bfit{\mu}_n(-N)\\ \bfit{\mu}_n(-N+1)\\ \ \vdots\\ \bfit{\mu}_n(N-1)\\ \bfit{\mu}_n(N)\end{array}\right] 
	= \left[\begin{array}{l} \mu_n(x,\,-N)\\ \mu_n(-x,\,-N)\\ \mu_n(x,\,-N+1)\\ \mu_n(-x,\,-N+1)\\ \ \vdots\\ \mu_n(x,\,N)\\ \mu_n(-x,\,N)\end{array}\right] \in [0,\,1]^{4N+2}.
\end{equation}
By Eqs.~\eqref{methods:eq:mu:nedge}, \eqref{methods:eq:mu:middle}, and \eqref{methods:eq:mu:pedge}, we obtain
\begin{equation}
	\bfit{\mu}_{n+1}
	=
	\left[
	\begin{array}{ccccccc}
		\Gamma P(-N) & \Gamma P(-N+1) & O & \cdots & \multicolumn{3}{c}{\raisebox{-25pt}[0pt][0pt]{\Huge $O$}}\\
		\Gamma Q(-N) & O & \Gamma P(-N+2) & \cdots &&&\\
		O & \Gamma Q(-N+1) & O & \cdots &&&\\
		\vdots & \vdots & \vdots & \ddots &&&\\
		\multicolumn{4}{c}{\raisebox{-20pt}[0pt][0pt]{\Huge $O$}} & O &\Gamma  P(N-1) & O\\
		&&&& \Gamma Q(N-2) & O & \Gamma P(N)\\
		&&&& O & \Gamma Q(N-1) & \Gamma Q(N)
	\end{array}
	\right]
	\bfit{\mu}_n.
\end{equation}
In the following, we denote the above matrix combining $\mu_n$ and $\mu_{n+1}$ by $M\in [0,\,1]^{(4N+2)\times(4N+2)}$, which is the transition probability matrix of the state transition on $(\sigma,\,i)\in\{\pm x\}\times\{0,\,\pm 1,\,\cdots,\,\pm(N-1),\,\pm N\}$.

Our intention is to obtain the limiting distribution $\lim_{n\to\infty} \mu_n$.  
It is well known that the limiting distribution of a Markov process coincides with its stationary distribution $\bfit{\pi}$, which satisfies
\begin{equation}
	\bfit{\pi} = M\bfit{\pi}.
\end{equation}
Thus, $\bfit{\pi}$ is a normalized eigenvector of $M$ associated with eigenvalue $1$.  
In the following, we derive the stationary distribution $\bfit{\pi}$ under the condition $p_A + p_B = 1$.

Let $p_A = p$ and $p_B = q$.  
When $p+q=1$ holds, $q(\tau,\,j)$ in Eq.~\eqref{methods:eq:q} is simply $q$, independent of the relation between the threshold $\sigma$ and the signal $i$.  
Therefore, the matrices $Q(i)$ and $P(i)$ in Eqs.~\eqref{methods:eq:Q} and \eqref{methods:eq:P} become
\begin{equation}
	Q(i) = qI_2,\qquad P(i) = pI_2
\end{equation}
for any threshold $i\in\{0,\,\pm 1,\,\cdots,\,\pm N\}$.
In this case, $M$ reduces to
\begin{equation}
	M =
	\left[
	\begin{array}{ccccccc}
		p\Gamma & p\Gamma & O & \cdots & \multicolumn{3}{c}{\raisebox{-25pt}[0pt][0pt]{\Huge $O$}}\\
		q\Gamma & O & p\Gamma & \cdots &&&\\
		O & q\Gamma & O & \cdots &&&\\
		\vdots & \vdots & \vdots & \ddots &&&\\
		\multicolumn{4}{c}{\raisebox{-20pt}[0pt][0pt]{\Huge $O$}} & O & p\Gamma & O\\
		&&&& q\Gamma & O & p\Gamma\\
		&&&& O & q\Gamma & q\Gamma
	\end{array}
	\right].
\end{equation}
Next, we define $\mathcal{P}\in [0,\,1]^{(2N+1)\times(2N+1)}$ by
\begin{equation}
	\mathcal{P} = 
	\left[\begin{array}{ccccccc}
		p & p & O & \cdots & \multicolumn{3}{c}{\raisebox{-25pt}[0pt][0pt]{\Huge $O$}}\\
		q & O & p & \cdots &&&\\
		O & q & O & \cdots &&&\\
		\vdots & \vdots & \vdots & \ddots &&&\\
		 \multicolumn{4}{c}{\raisebox{-20pt}[0pt][0pt]{\Huge $O$}} & O & p & O\\
		 &&&& q & O & p\\
		 &&&& O & q & q
		 \end{array}\right].
\end{equation}
Then, $M$ can be decomposed as a Kronecker product:
\begin{equation}
	M = 	\left[\begin{array}{ccccccc}
		p & p & O & \cdots & \multicolumn{3}{c}{\raisebox{-25pt}[0pt][0pt]{\Huge $O$}}\\
		q & O & p & \cdots &&&\\
		O & q & O & \cdots &&&\\
		\vdots & \vdots & \vdots & \ddots &&&\\
		 \multicolumn{4}{c}{\raisebox{-20pt}[0pt][0pt]{\Huge $O$}} & O & p & O\\
		 &&&& q & O & p\\
		 &&&& O & q & q
		 \end{array}\right]\otimes \Gamma = \mathcal{P}\otimes \Gamma.
\end{equation}
By the property of diagonalizable matrices represented by Kronecker products, eigenvalues of $M$ are obtained by calculating those of $\mathcal{P}$ and $\Gamma$;
that is, if $\alpha$ and $\beta$ are eigenvalues of $\mathcal{P}$ and $\Gamma$, respectively, then 
their product $\alpha\beta$ is an eigenvalue of $M = \mathcal{P}\otimes \Gamma$.
Moreover, if $\bfit{u}\in \mathbb{R}^{2N+1}$ and $\bfit{v}\in \mathbb{R}^2$ are eigenvectors of $\mathcal{P}$ and $\Gamma$ belonging to $\varepsilon$ and $\varepsilon'$, respectively, then $\bfit{u}\otimes \bfit{v}$ is an eigenvector of $M$ belonging to $\varepsilon\varepsilon'$.
Therefore, our target is obtained by pursuing the eigenvector associated with eigenvalue $1$ for $\mathcal{P}$ and $\Gamma$, respectively.
Solving $\bfit{u}=\mathcal{P}\bfit{u}$ and $\bfit{v}=\Gamma\bfit{v}$ yields
\begin{equation}
	\bfit{u} = c\left[\begin{array}{c} 1\\ \vdots\\ (q/p)^{N+i}\\ \vdots\\ (q/p)^{2N}\end{array}\right]
	\quad\text{and}\quad
	\bfit{v} = c'\twovec{1}{1},
\end{equation}
where $c$ and $c'$ are constants, and $(q/p)^{N+i}$ appears in the $(N+i+1)$-th row entry for $i\in\{0,\,\pm 1,\,\cdots,\,\pm N\}$.
Thus, an eigenvector of $M$ belonging to eigenvalue $1$ is
\begin{equation}
	\bfit{u}\otimes \bfit{v} = cc' \left[\begin{array}{c} 1\\ \vdots\\ (q/p)^{N+i}\\ \vdots\\ (q/p)^{2N}\end{array}\right]\otimes \twovec{1}{1} = C \left[\begin{array}{c} 1\\ 1\\ \vdots\\ (q/p)^{N+i}\\ (q/p)^{N+i} \\ \vdots\\ (q/p)^{2N}\\ (q/p)^{2N}\end{array}\right]
\end{equation}
with $C = cc'$. Here,$(q/p)^{N+i}$ is in the $\{2(N+i+1)-1\}$-th and $2(N+i+1)$-th row entries in the resultant form.
Normalizing this vector, we obtain the stationary distribution
\begin{equation}\label{proof:eq:pi}
	\bfit{\pi} = \fraction{1}{\displaystyle 2\sum_{k=1}^{2N+1} (q/p)^{k-1}} \left[\begin{array}{c} 1\\ 1\\ \vdots\\ (q/p)^{N+i}\\ (q/p)^{N+i} \\ \vdots\\ (q/p)^{2N}\\ (q/p)^{2N}\end{array}\right] 
	= \fraction{p-q}{2(p^{2N+1} -q^{2N+1})}\left[\begin{array}{c} p^{2N}\\ p^{2N}\\ \vdots\\ p^{N-i}q^{N+i}\\ p^{N-i}q^{N+i}\\ \vdots\\ q^{2N}\\ q^{2N}\end{array}\right].
\end{equation}
Hence, the limiting distribution of the Markov process is $\bfit{\pi}$, describing the time-series-based decision making driven by the two-valued signal $s_n$.  
In particular, the probability of the pair $(\pm x,\,i)$ is
\begin{equation}\label{proof:eq:pientry}
	\pi (x,\,i) = \pi(-x,\,i) = \fraction{(p-q)p^N q^N}{2(p^{2N+1} -q^{2N+1})}\left(\fraction{q}{p}\right)^i.
\end{equation}
Applying this formula to the limiting value of $\CDR_n$ shown in Eq.~\eqref{modeling:eq:cdr2} as $n\to \infty$, we obtain
\begin{equation}\begin{split}
	\CDR_\infty &= \lim_{n\to\infty}\CDR_n\\
	&= \sum_{i = -N}^{\lfloor x \rfloor} \pi (x,\,i) + \sum_{i = -N}^{-\lfloor x \rfloor-1} \pi (-x,\,i)\\
	&= \fraction{(p-q)p^N q^N}{2(p^{2N+1} -q^{2N+1})}\left\{\sum_{i = -N}^{\lfloor x \rfloor} \left(\fraction{q}{p}\right)^i
					 + \sum_{i = -N}^{-\lfloor x \rfloor-1} \left(\fraction{q}{p}\right)^i \right\}\\
		&= \fraction{(p-q)p^N q^N}{2(p^{2N+1} -q^{2N+1})}\cdot \fraction{p^{N+1}}{q^N (p-q)} \left\{2 -\left(\fraction{q}{p}\right)^{N+\lfloor x \rfloor+1} -\left(\fraction{q}{p}\right)^{N-\lfloor x \rfloor}\right\}\\
		&= \fraction{2p^{2N+1} -p^{N-\lfloor x \rfloor}q^{N+\lfloor x \rfloor+1} -p^{N+\lfloor x \rfloor+1}q^{N-\lfloor x \rfloor}}{2(p^{2N+1} -q^{2N+1})}.
\end{split}\end{equation}

\vspace{\baselineskip}
\noindent
\textbf{Acknowledgments:} 
This work was supported in part by JSPS KAKENHI (JP22H05195, JP22H05197, JP23KJ0384, JP25K21294, JP25H01129) and JST CREST (JPMJCR24R2).\par

\vspace{\baselineskip}\noindent
\textbf{Data availability statement:} The data that support the findings of this study are available from the corresponding author upon reasonable request.

\vspace{\baselineskip}
\noindent
\textbf{Competing interests}: The authors declare no competing interests.

\bibliographystyle{tex_elements/naturemag}
\bibliography{tex_elements/tybib}

@article{appeltant2011information,
  title={Information processing using a single dynamical node as complex system},
  author={Appeltant, Lennert and Soriano, Miguel Cornelles and Van der Sande, Guy and Danckaert, Jan and Massar, Serge and Dambre, Joni and Schrauwen, Benjamin and Mirasso, Claudio R and Fischer, Ingo},
  journal={Nature Communications},
  volume={2},
  number={1},
  pages={468},
  year={2011},
  publisher={Nature Publishing Group UK London}
}

@article{argyris2005chaos,
  title={Chaos-based communications at high bit rates using commercial fibre-optic links},
  author={Argyris, Apostolos and Syvridis, Dimitris and Larger, Laurent and Annovazzi-Lodi, Valerio and Colet, Pere and Fischer, Ingo and Garcia-Ojalvo, Jordi and Mirasso, Claudio R and Pesquera, Luis and Shore, K Alan},
  journal={Nature},
  volume={438},
  number={7066},
  pages={343--346},
  year={2005},
  publisher={Nature Publishing Group UK London}
}

@article{argyris2010implementation,
  title={Implementation of 140 $\mathrm{Gb/s}$ true random bit generator based on a chaotic photonic integrated circuit},
  author={Argyris, Apostolos and Deligiannidis, Stavros and Pikasis, Evangelos and Bogris, Adonis and Syvridis, Dimitris},
  journal={Optics Express},
  volume={18},
  number={18},
  pages={18763--18768},
  year={2010},
  publisher={OSA}
}

@article{auer2002finite,
  title={Finite-time analysis of the multiarmed bandit problem},
  author={Auer, Peter and Cesa-Bianchi, Nicolo and Fischer, Paul},
  journal={Machine Learning},
  volume={47},
  number={2},
  pages={235--256},
  year={2002},
  publisher={Springer}
}

@article{colet1994digital,
  title={Digital communication with synchronized chaotic lasers},
  author={Colet, Pere and Roy, Rajarshi},
  journal={Optics Letters},
  volume={19},
  number={24},
  pages={2056--2058},
  year={1994},
  publisher={Optical Society of America}
}

@article{cuevas2024solving,
  title={Solving multi-armed bandit problems using a chaotic microresonator comb},
  author={Cuevas, Jonathan and Iwami, Ryugo and Uchida, Atsushi and Minoshima, Kaoru and Kuse, Naoya},
  journal={APL Photonics},
  volume={9},
  number={3},
  pages={036112},
  year={2024},
  publisher={AIP Publishing}
}

@article{daw2006cortical,
  title={Cortical substrates for exploratory decisions in humans},
  author={Daw, Nathaniel D and O'doherty, John P and Dayan, Peter and Seymour, Ben and Dolan, Raymond J},
  journal={Nature},
  volume={441},
  number={7095},
  pages={876--879},
  year={2006},
  publisher={Nature Publishing Group UK London}
}

@article{duport2012all,
  title={All-optical reservoir computing},
  author={Duport, Fran{\c{c}}ois and Schneider, Bendix and Smerieri, Anteo and Haelterman, Marc and Massar, Serge},
  journal={Optics Express},
  volume={20},
  number={20},
  pages={22783--22795},
  year={2012},
  publisher={Optical Society of America}
}

@article{hasegawa2016improving,
  title={Improving performance of heuristic algorithms by Lebesgue spectrum filter},
  author={Hasegawa, Mikio},
  journal={IEICE Transactions on Communications},
  volume={99.B},
  number={11},
  pages={2256--2262},
  year={2016},
  publisher={The Institute of Electronics, Information and Communication Engineers}
}

@article{hinrichs2020persistence,
  title={Persistence of one-dimensional $\mathrm{AR}(1)$-sequences},
  author={Hinrichs, G{\"u}nter and Kolb, Martin and Wachtel, Vitali},
  journal={Journal of Theoretical Probability},
  volume={33},
  number={1},
  pages={65--102},
  year={2020},
  publisher={Springer}
}

@article{homma2019chip,
  title={On-chip photonic decision maker using spontaneous mode switching in a ring laser},
  author={Homma, Ryutaro and Kochi, Satoshi and Niiyama, Tomoaki and Mihana, Takatomo and Mitsui, Yusuke and Kanno, Kazutaka and Uchida, Atsushi and Naruse, Makoto and Sunada, Satoshi},
  journal={Scientific Reports},
  volume={9},
  pages={9429},
  year={2019},
  publisher={Nature Publishing Group UK London},
  doi={10.1038/s41598-019-45754-3}
}

@article{iwami2022controlling,
  title={Controlling chaotic itinerancy in laser dynamics for reinforcement learning},
  author={Iwami, Ryugo and Mihana, Takatomo and Kanno, Kazutaka and Sunada, Satoshi and Naruse, Makoto and Uchida, Atsushi},
  journal={Science Advances},
  volume={8},
  number={49},
  pages={eabn8325},
  year={2022},
  publisher={American Association for the Advancement of Science}
}

@book{jahns2014optical,
  title={Optical Computing Hardware: Optical Computing},
  author={Jahns, J{\"u}rgen and Lee, Sing H},
  year={2014},
  publisher={Academic press}
}

@article{kanno2020adaptive,
  title={Adaptive model selection in photonic reservoir computing by reinforcement learning},
  author={Kanno, Kazutaka and Naruse, Makoto and Uchida, Atsushi},
  journal={Scientific Reports},
  volume={10},
  pages={10062},
  year={2020},
  publisher={Nature Publishing Group UK London}
}

@article{kanter2010optical,
  title={An optical ultrafast random bit generator},
  author={Kanter, Ido and Aviad, Yaara and Reidler, Igor and Cohen, Elad and Rosenbluh, Michael},
  journal={Nature Photonics},
  volume={4},
  number={1},
  pages={58--61},
  year={2010},
  publisher={Nature Publishing Group UK London}
}

@article{kim2010tug,
  title={Tug-of-war model for the two-bandit problem: Nonlocally-correlated parallel exploration via resource conservation},
  author={Kim, Song-Ju and Aono, Masashi and Hara, Masahiko},
  journal={BioSystems},
  volume={101},
  number={1},
  pages={29--36},
  year={2010},
  publisher={Elsevier},
  doi={10.1016/j.biosystems.2010.04.002}
}

@article{kim2015efficient,
  title={Efficient decision-making by volume-conserving physical object},
  author={Kim, Song-Ju and Aono, Masashi and Nameda, Etsushi},
  journal={New Journal of Physics},
  volume={17},
  number={8},
  pages={083023},
  year={2015},
  publisher={IOP Publishing}
}

@article{kitayama2019novel,
  title={Novel frontier of photonics for data processing—Photonic accelerator},
  author={Kitayama, Kenichi and Notomi, Masaya and Naruse, Makoto and Inoue, Koji and Kawakami, Satoshi and Uchida, Atsushi},
  journal={APL Photonics},
  volume={4},
  number={9},
  pages={090901},
  year={2019},
  publisher={AIP Publishing}
}

@article{kohda2003pursley,
  title={Pursley's aperiodic cross-correlation functions revisited},
  author={Kohda, Tohru and Fujisaki, Hiroshi},
  journal={IEEE Transactions on Circuits and Systems I: Fundamental Theory and Applications},
  volume={50},
  number={6},
  pages={800--805},
  year={2003},
  publisher={IEEE}
}

@inproceedings{lai2008medium,
  title={Medium access in cognitive radio networks: A competitive multi-armed bandit framework},
  author={Lai, Lifeng and Gamal, Hesham El and Poor, H Vincent},
  booktitle = {Proceedings of IEEE 42nd Asilomar Conference on Signals, System and Computer},
  pages = {98--102},
  year = {2008}
}

@article{march1991exploration,
  title={Exploration and exploitation in organizational learning},
  author={March, James G},
  journal={Organization Science},
  volume={2},
  number={1},
  pages={71--87},
  year={1991},
  publisher={INFORMS},
  doi={10.1287/orsc.2.1.71}
}

@article{mihana2018memory,
  title={Memory effect on adaptive decision making with a chaotic semiconductor laser},
  author={Mihana, Takatomo and Terashima, Yuta and Naruse, Makoto and Kim, Song-Ju and Uchida, Atsushi},
  journal={Complexity},
  volume={2018},
  pages={4318127},
  year={2018},
  publisher={Wiley Online Library}
}

@article{morijiri2023parallel,
  title={Parallel photonic accelerator for decision making using optical spatiotemporal chaos},
  author={Morijiri, Kensei and Takehana, Kento and Mihana, Takatomo and Kanno, Kazutaka and Naruse, Makoto and Uchida, Atsushi},
  journal={Optica},
  volume={10},
  number={3},
  pages={339--348},
  year={2023},
  publisher={Optica Publishing Group},
  doi={10.1364/OPTICA.477433}
}

@article{narimatsu2025study,
doi = {10.1088/1742-5468/ae0560},
year = {2025},
publisher = {IOP Publishing},
volume = {2025},
number = {9},
pages = {093407},
author = {Narimatsu, Akihiro and Yamagami, Tomoki},
title = {A study of the antlion random walk},
journal = {Journal of Statistical Mechanics: Theory and Experiment}
}

@article{nakamura2005improvements,
  title={Improvements to the linear programming based scheduling of web advertisements},
  author={Nakamura, Atsuyoshi and Abe, Naoki},
  journal={Electronic Commerce Research},
  volume={5},
  pages={75--98},
  year={2005},
  publisher={Springer}
}

@article{naruse2017ultrafast,
  title={Ultrafast photonic reinforcement learning based on laser chaos},
  author={Naruse, Makoto and Terashima, Yuta and Uchida, Atsushi and Kim, Song-Ju},
  journal={Scientific Reports},
  volume={7},
  pages={8772},
  year={2017},
  publisher={Nature Publishing Group UK London}
}

@article{naruse2018scalable,
  title={Scalable photonic reinforcement learning by time-division multiplexing of laser chaos},
  author={Naruse, Makoto and Mihana, Takatomo and Hori, Hirokazu and Saigo, Hayato and Okamura, Kazuya and Hasegawa, Mikio and Uchida, Atsushi},
  journal={Scientific Reports},
  volume={8},
  pages={10890},
  year={2018},
  publisher={Nature Publishing Group UK London}
}

@book{ohtsubo2012semiconductor,
  title={Semiconductor Lasers: Stability, Instability and Chaos, 4th Ed.},
  author={Ohtsubo, Junji},
  year={2017},
  publisher={Springer},
  doi={10.1007/978-3-319-56138-7}
}

@article{okada2021analysis,
  title={Analysis on effectiveness of surrogate data-based laser chaos decision maker},
  author={Okada, Norihiro and Hasegawa, Mikio and Chauvet, Nicolas and Li, Aohan and Naruse, Makoto},
  journal={Complexity},
  volume={2021},
  pages={8877660},
  year={2021},
  publisher={Hindawi Limited}
}

@article{okada2022theory,
  title={Theory of acceleration of decision-making by correlated time sequences},
  author={Okada, Norihiro and Yamagami, Tomoki and Chauvet, Nicolas and Ito, Yusuke and Hasegawa, Mikio and Naruse, Makoto},
  journal={Complexity},
  volume={2022},
  pages={5205580},
  year={2022},
  publisher={Hindawi Limited}
}

@article{peng2021photonic,
  title={Photonic decision-making for arbitrary-number-armed bandit problem utilizing parallel chaos generation},
  author={Peng, Jiafa and Jiang, Ning and Zhao, Anke and Liu, Shiqin and Zhang, Yiqun and Qiu, Kun and Zhang, Qianwu},
  journal={Optics Express},
  volume={29},
  number={16},
  pages={25290--25301},
  year={2021},
  publisher={Optical Society of America}
}

@article{robbins1952some,
  title={Some aspects of the sequential design of experiments},
  author={Robbins, Herbert},
  journal={Bulletin of the American Mathematical Society},
  volume={58},
  pages={527--535},
  year={1952},
  url={https://www.ams.org/journals/bull/1952-58-05/S0002-9904-1952-09620-8/}
}

@article{scheuer2006giant,
  title={Giant fiber lasers: A new paradigm for secure key distribution},
  author={Scheuer, Jacob and Yariv, Amnon},
  journal={Physical Review Letters},
  volume={97},
  number={14},
  pages={140502},
  year={2006},
  publisher={APS}
}

@article{shen2023harnessing,
  title={Harnessing microcomb-based parallel chaos for random number generation and optical decision making},
  author={Shen, Bitao and Shu, Haowen and Xie, Weiqiang and Chen, Ruixuan and Liu, Zhi and Ge, Zhangfeng and Zhang, Xuguang and Wang, Yimeng and Zhang, Yunhao and Cheng, Buwen and others},
  journal={Nature Communications},
  volume={14},
  number={1},
  pages={4590},
  year={2023},
  publisher={Nature Publishing Group UK London}
}

@article{shastri2021photonics,
  title={Photonics for artificial intelligence and neuromorphic computing},
  author={Shastri, Bhavin J and Tait, Alexander N and Ferreira de Lima, Thomas and Pernice, Wolfram HP and Bhaskaran, Harish and Wright, C David and Prucnal, Paul R},
  journal={Nature Photonics},
  volume={15},
  number={2},
  pages={102--114},
  year={2021},
  publisher={Nature Publishing Group UK London}
}

@inproceedings{solomyak2004notes,
  title={Notes on Bernoulli convolutions},
  author={Solomyak, Boris},
  booktitle={Proceedings of Symposia in Pure Mathematics},
  volume={1},
  pages={207--230},
  year={2004}
}

@article{soriano2013complex,
  title={Complex photonics: Dynamics and applications of delay-coupled semiconductors lasers},
  author={Soriano, Miguel C and Garc{\'\i}a-Ojalvo, Jordi and Mirasso, Claudio R and Fischer, Ingo},
  journal={Reviews of Modern Physics},
  volume={85},
  number={1},
  pages={421--470},
  year={2013},
  publisher={APS},
  doi={10.1103/RevModPhys.85.421}
}

@book{sutton2018reinforcement,
  title={Reinforcement Learning: An Introduction, 2nd Ed.},
  author={Sutton, Richard S and Barto, Andrew G},
  year={2018},
  publisher={MIT press},
  url={http://incompleteideas.net/book/the-book-2nd.html}
}

@article{takeuchi2020dynamic,
  title={Dynamic channel selection in wireless communications via a multi-armed bandit algorithm using laser chaos time series},
  author={Takeuchi, Shungo and Hasegawa, Mikio and Kanno, Kazutaka and Uchida, Atsushi and Chauvet, Nicolas and Naruse, Makoto},
  journal={Scientific Reports},
  volume={10},
  pages={1574},
  year={2020},
  publisher={Nature Publishing Group UK London},
  doi={10.1038/s41598-020-58541-2}
}

@article{thompson1933likelihood,
  title={On the likelihood that one unknown probability exceeds another in view of the evidence of two samples},
  author={Thompson, William R},
  journal={Biometrika},
  volume={25},
  number={3-4},
  pages={285--294},
  year={1933},
  publisher={Oxford University Press}
}

@article{uchida2008fast,
  title={Fast physical random bit generation with chaotic semiconductor lasers},
  author={Uchida, Atsushi and Amano, Kazuya and Inoue, Masaki and Hirano, Kunihito and Naito, Sunao and Someya, Hiroyuki and Oowada, Isao and Kurashige, Takayuki and Shiki, Masaru and Yoshimori, Shigeru and Yoshimura, Kazuyuki and Davis, Peter},
  journal={Nature Photonics},
  volume={2},
  number={12},
  pages={728--732},
  year={2008},
  publisher={Nature Publishing Group}
}

@article{yamagami2023bandit,
  title={Bandit algorithm driven by a classical random walk and a quantum walk},
  author={Yamagami, Tomoki and Segawa, Etsuo and Mihana, Takatomo and R{\"o}hm, Andr{\'e} and Horisaki, Ryoichi and Naruse, Makoto},
  journal={Entropy},
  volume={25},
  number={6},
  pages={843},
  year={2023},
  publisher={MDPI},
  doi={10.3390/e25060843}
}

@article{yasudo2025photonic,
  title={Photonic Ising machine using parallel bandit architecture based on laser chaos},
  author={Yasudo, Ryota and Koibuchi, Michihiro and Hasegawa, Mikio},
  journal={IEEE Access},
  volume={13},
  pages={211981--211995},
  year={2025},
  publisher={IEEE}
}

@article{yoshimura2012secure,
  title={Secure key distribution using correlated randomness in lasers driven by common random light},
  author={Yoshimura, Kazuyuki and Muramatsu, Jun and Davis, Peter and Harayama, Takahisa and Okumura, Haruka and Morikatsu, Shinichiro and Aida, Hiroki and Uchida, Atsushi},
  journal={Physical Review Letters},
  volume={108},
  number={7},
  pages={070602},
  year={2012},
  publisher={APS}
}
\end{document}